\documentclass[sigconf]{acmart}

\usepackage{latexsym}

\usepackage{amssymb}
\usepackage{amsmath}
\usepackage{amsthm}
\usepackage{booktabs}
\usepackage{enumitem}
\usepackage{graphicx}
\usepackage{color}

\usepackage{algorithm}
\usepackage{algpseudocode}
\usepackage{multirow}
\usepackage{subcaption}

\AtBeginDocument{%
  }
    
\newtheorem{definition}{Definition}

\copyrightyear{2025}
\acmYear{2025}
\setcopyright{acmlicensed}\acmConference[MM '25]{Proceedings of the 33rd ACM International Conference on Multimedia}{October 27--31, 2025}{Dublin, Ireland}
\acmBooktitle{Proceedings of the 33rd ACM International Conference on Multimedia (MM '25), October 27--31, 2025, Dublin, Ireland}
\acmDOI{10.1145/3746027.3754842}
\acmISBN{979-8-4007-2035-2/2025/10}

\begin{document}

\title{DisMS-TS: Eliminating Redundant Multi-Scale Features for Time Series Classification}

\author{Zhipeng Liu}
\affiliation{%
  \institution{School of Software, Northeastern University}
  \city{Shenyang}
  \country{China}}
\email{2310543@stu.neu.edu.cn}

\author{Peibo Duan$^*$}
\affiliation{%
  \institution{School of Software, Northeastern University}
  \city{Shenyang}
  \country{China}}
\email{duanpeibo@swc.neu.edu.cn}\thanks{Peibo Duan and Binwu Wang are Corresponding authors.}

\author{Binwu Wang$^*$}
\affiliation{%
  \institution{School of Software, University of Science and Technology of China}
  \city{Hefei}
  \country{China}}
\email{wbw2024@ustc.edu.cn
}

\author{Xuan Tang}
\affiliation{%
  \institution{School of Software, Northeastern University}
  \city{Shenyang}
  \country{China}}
\email{2471477@stu.neu.edu.cn}

\author{Qi Chu}
\affiliation{%
  \institution{School of Software, Northeastern University}
  \city{Shenyang}
  \country{China}}
\email{chuqineu@gmail.com}

\author{Changsheng Zhang}
\affiliation{%
  \institution{School of Software, Northeastern University}
  \city{Shenyang}
  \country{China}}
\email{zhangchangsheng@mail.neu.edu.cn}

\author{Yongsheng Huang}
\affiliation{%
  \institution{School of Software, Northeastern University}
  \city{Shenyang}
  \country{China}}
\email{2371447@stu.neu.edu.cn}

\author{Bin Zhang}
\affiliation{%
  \institution{School of Software, Northeastern University}
  \city{Shenyang}
  \country{China}}
\email{zhangbin@mail.neu.edu.cn}

\renewcommand{\shortauthors}{Zhipeng Liu et al.}

\begin{abstract}
  Real-world time series typically exhibit complex temporal variations, making the time series classification task notably challenging. Recent advancements have demonstrated the potential of multi-scale analysis approaches, which provide an effective solution for capturing these complex temporal patterns. However, existing multi-scale analysis-based time series prediction methods fail to eliminate redundant scale-shared features across multi-scale time series, resulting in the model over- or under-focusing on scale-shared features. To address this issue, we propose a novel end-to-end \textbf{Dis}entangled \textbf{M}ulti-\textbf{S}cale framework for \textbf{T}ime \textbf{S}eries classification (\textbf{DisMS-TS}). The core idea of DisMS-TS is to eliminate redundant shared features in multi-scale time series, thereby improving prediction performance. Specifically, we propose a temporal disentanglement module to capture scale-shared and scale-specific temporal representations, respectively. Subsequently, to effectively learn both scale-shared and scale-specific temporal representations, we introduce two regularization terms that ensure the consistency of scale-shared representations and the disparity of scale-specific representations across all temporal scales. Extensive experiments conducted on multiple datasets validate the superiority of DisMS-TS over its competitive baselines, with the accuracy improvement up to 9.71\%. The source code is publicly available on GitHub\footnote{https://github.com/ZhipengLiu75/DisMS-TS}.
\end{abstract}

\ccsdesc[500]{Mathematics of computing~Time series analysis}

\keywords{Time Series, Multi-scale Analysis, Feature Disentanglement}

\maketitle

\section{Introduction}
\label{intro}

    Time series have garnered increasing attention due to their widespread applications in various real-world scenarios \cite{huang2025exploiting,miao2025parameter}, such as healthcare \cite{morid2023time}, financial investment \cite{liu2025dynamic,liu2025attributed}, and traffic planning \cite{DBLP:journals/corr/abs-2401-10134,zhang2025strap}. Time series classification plays a pivotal role in addressing these challenges by categorizing time series into predefined classes based on their historical observations \cite{wang2024survey,jin2024survey}. However, the observed time series often exhibit complex temporal dynamics due to the inherent volatility of real-world systems, making the prediction task notably challenging.

	\begin{figure}[t]
		\centering
        \includegraphics[width=.48\textwidth]{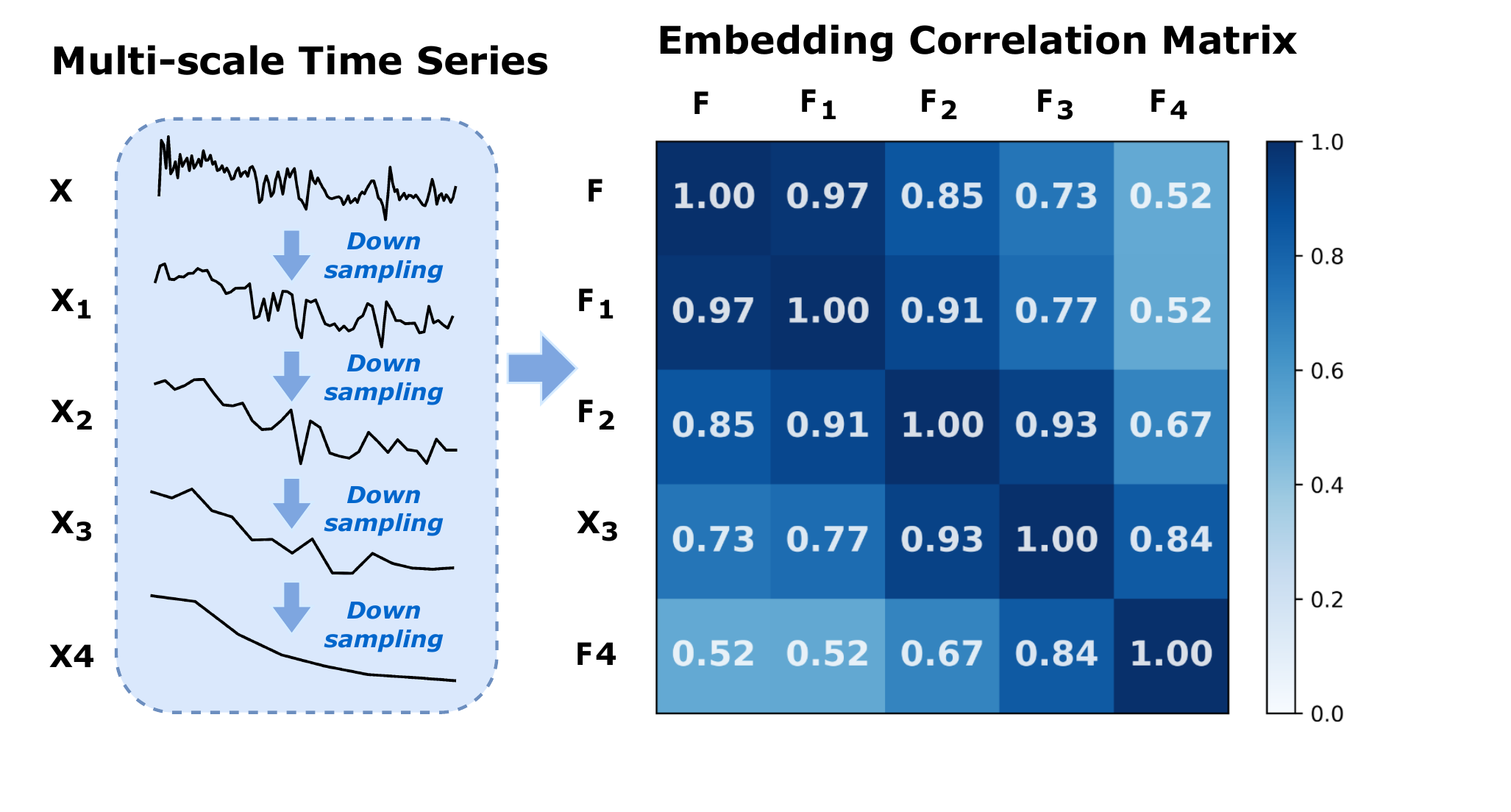}
		\caption{Correlation analysis of multi-scale temporal features. The left represents time series at five different scales, while the right displays the corresponding correlation matrix. A higher correlation score indicates greater redundancy.
        }
		\label{intropearson}
	\end{figure}

    In recent years, deep learning techniques have demonstrated remarkable progress in time series prediction. For instance, recurrent neural networks (RNNs) and their variants, such as long short-term memory (LSTM) and gated recurrent units (GRUs) \cite{stankeviciute2021conformal,zhu2023long}, along with convolutional neural networks (CNNs) \cite{lee2021DPT} and Transformers \cite{zhang2023crossformer}, are extensively utilized as backbones, owing to their exceptional performance in addressing time series data. Moreover, since time series exhibit distinct temporal variations at different sampling scales, multi-scale analysis approaches are widely integrated into these backbones to capture more intricate temporal patterns \cite{chen_2021_multiscale_Neural_Networks,guo2023icassp,zhong2023mlpmixer,MICN,cheng2023www}. For example, daily stock data reflects short-term fluctuations, whereas seasonal stock data demonstrates long-term fundamentals \cite{teng_2022_multistock}.  The vast majority of multi-scale analysis methods follow a paradigm in which the time series is sampled at multiple scales, with specialized analyses of temporal patterns conducted at each scale. The results of these analyses, i.e., temporal embeddings of individual scales, are then fused for prediction using scale-wise feature fusion techniques, such as attention mechanisms. For instance, MAGNN \cite{chen_2023_MAGNN} is designed to assign weights to embeddings of different temporal scales for prediction, and TimeMixer \cite{wang_2024_Timemixer} includes bottom-up and top-down mixing approaches for fusing multi-scale seasonal and trend features, respectively, for subsequent prediction. 


    Although these works address the multi-scale properties of time series, they overlook the impact of redundant features across different scales, which may result in the model over- or under-focusing on scale-shared features. Since the multi-scale time series are sampled from the original observations, they contain some temporal variations that may overlap, which can lead to redundancy in the multi-scale features. To better illustrate, as shown in Figure \ref{intropearson}, we utilize the real-world Human Activity Recognition (HAR) dataset as a case study for preliminary analysis. First, the original observation data of HAR is downsampled into four scales via widely employed average pooling \cite{wang_2024_Timemixer}. The right side of Figure 1 illustrates the correlations among temporal embeddings at five scales derived from the well-known GRU. It can be observed that features across all temporal scales exhibit a high degree of correlation, which indicates the existence of a large number of redundant features shared by multi-scale time series (please refer to Sec.\ref{method} for the definition of feature redundancy). However, due to the entanglement of scale-shared and scale-specific features at individual temporal scales, scale-wise feature fusion approaches cannot eliminate the redundant scale-shared features. This may result in the model over- or under-focusing on scale-shared features (please refer to Sec.\ref{motivation} for theoretical analysis), potentially jeopardizing prediction performance.

    To address the aforementioned problems, we propose a novel end-to-end \textbf{Dis}entangled \textbf{M}ulti-\textbf{S}cale framework for \textbf{T}ime \textbf{S}eries classification, named \textbf{DisMS-TS}. The core idea of DisMS-TS is to eliminate redundant features across all temporal scales while preserving the unique features of each scale. Specifically, we first conduct multi-layer downsampling on the historical observation series, generating variant series at different temporal scales for multi-view analysis. After that, these multi-scale time series are treated independently for capturing temporal representations and subsequently fed into the temporal disentanglement modules, where they are processed separately to disentangle scale-shared and scale-specific representations. Finally, we design two regularization terms, i.e., similarity and disparity loss, to maximize the similarity of scale-shared representations across different time scales and the disparity between scale-specific representations, respectively. In this way, our model can eliminate redundant scale-shared features while leveraging both scale-shared and scale-specific representations to achieve more accurate predictions. In summary, our contributions can be summarized as follows:
    
	\begin{itemize}
	\item We make a key observation that there are significant redundant features across different temporal scales and provide a theoretical analysis of the limitations inherent in existing multi-scale analysis-based methods. To the best of our knowledge, this is the first study to consider eliminating redundant features in multi-scale time series.
        
        \item We propose a novel disentanglement module to capture both scale-shared and scale-specific temporal representations across different scales. Moreover, we propose two regularization terms to ensure the consistency of scale-shared representations and the disparity of scale-specific representations across different temporal scales, respectively.

        
        
        

        \item We conduct extensive experiments on multiple datasets and compare our proposed DisMS-TS with various classic and state-of-the-art methods. The experimental results, supported by statistical tests, demonstrate that our proposed DisMS-TS outperforms all its counterparts. 
	\end{itemize}

\section{Related work}	
\label{relaredwork}

\subsection{Time Series Analysis}

Time series analysis is a fundamental research topic with broad applications in downstream tasks \cite{qiu2024tfb,qiu2024duet,wang2023drift,lin2024sparsetsf,miao2024unified,ye2025non}, such as time series classification. Recently, with advancements in computational power, deep neural networks have garnered widespread attention in the research community. For example, CNNs employ temporal convolutions to capture local dependencies within subsequences \cite{sen2019think,liu2022scinet}. Some studies have introduced dynamic temporal pooling to enhance the representational capacity of CNNs for time series forecasting. Another well-established architecture is the RNN and its variants \cite{huang2019dsanet,kieu2022anomaly}. Certain research, like LSTNet, integrates temporal attention mechanism-based forecasting into RNNs to enhance predictions \cite{qin2017dual,lai2018lstnet}. TS-TCC \cite{eldele2021time} employs data augmentation to convert time series into two views and proposes a temporal contrastive module that learns robust representations through cross-view prediction . Transformers, based on the encoder-decoder architecture, capture diverse temporal dependencies by introducing customized attention mechanisms \cite{wu2021autoformer,zhou2022fedformer,zhang2023crossformer}. For instance, PatchTST \cite{patchTST} integrates patch and transformer architectures, combining channel independence to efficiently process multivariate time series. To model the correlation among channel variables, GNNs are incorporated into the learning of temporal representations \cite{li2021hierarchical,wang2023pattern,cheng2023weakly,liu2025distillation,huang2023crossgnn}. Inspired by the remarkable advancements of LLMs in the field of Natural Language Processing, LLMs have also been introduced to time series analysis tasks, demonstrating powerful zero-shot capabilities \cite{jin2023time,zhou2023one,liu2025timekd,liu2025st,liu2025timecma}.

Despite these methodologies have greatly advanced temporal analysis, they overlook the multi-scale characteristics of time series data, limiting the model's ability to capture more comprehensive information, such as trend patterns, ultimately affecting prediction performance.

\begin{figure*}[t]
  \centering
  \includegraphics[scale=0.54]{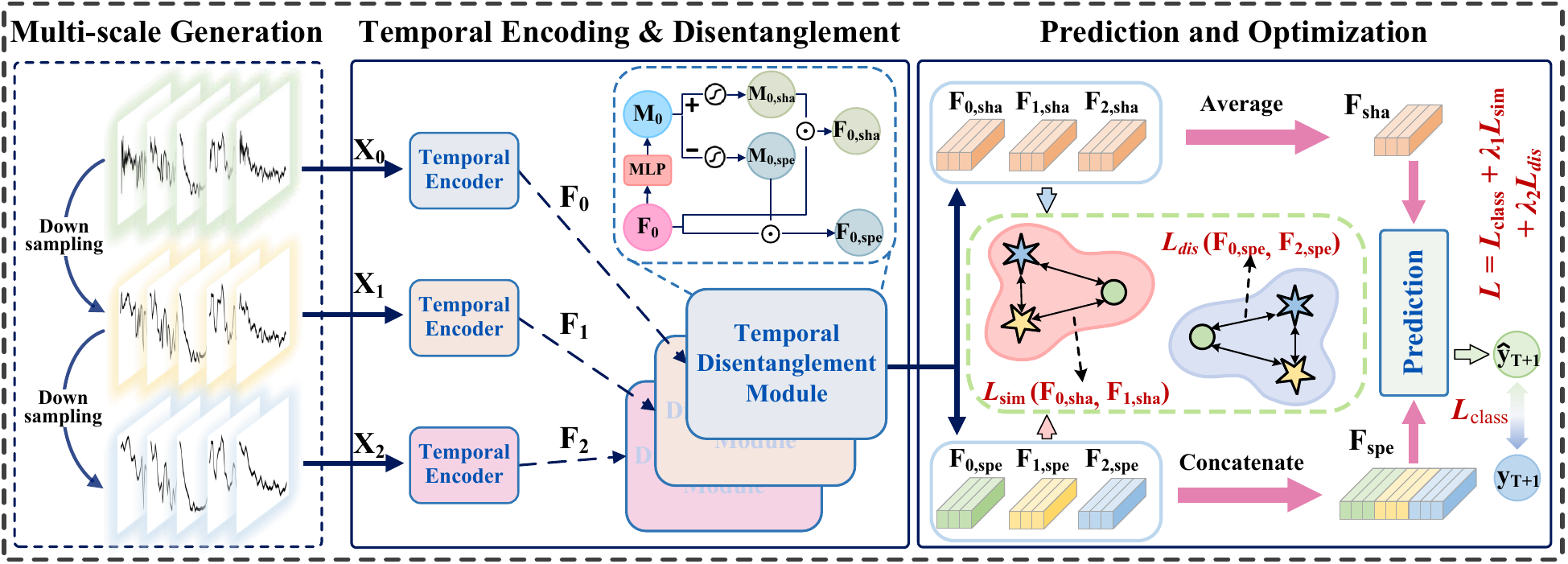}
  \caption{The framework of DisMS-TS. $S$ is assumed to be 2 for the sake of visualization.}
  \label{fig2}
\end{figure*}

\subsection{Multi-scale Analysis in Time Series}
Multi-scale analysis has proven to be effective for representation learning in fields such as computer vision and multi-modal learning \cite{wang2021pyramid,hu2020iterative}, which has led researchers to apply it in the time series domain. These methods follow a framework in which the input time series is first sampled into multi-scale time series, then fed into specialized temporal encoders to capture multi-scale temporal dependencies, and finally, scale-wise fusion techniques are applied for prediction. For instance, Pyraformer \cite{liu2022pyraformer} introduces pyramid attention to extract features at different temporal resolutions. MAGNN constructs a dynamic multi-scale adaptive graph to model the inter-variable relationships in multivariate time series for prediction \cite{chen_2023_MAGNN}. Chen et al. proposed Pathformer, which captures both local intra-patch and global inter-patch dependencies in multi-scale time series through a dual attention mechanism. Drawing inspiration from the multi-scale transformer language model \cite{subramanian2020multi}, Wang et al. proposed top-down and bottom-up mixing modules to fuse multi-scale trend and seasonal features, respectively \cite{wang_2024_Timemixer}. These methods can analyze time series from different temporal perspectives.

However, these methods overlook the impact of redundant scale-shared features across different temporal scales on prediction performance. Although a few methods select the optimal $K$ scale features for prediction \cite{chen2021scaleselect1,chen2024pathformer}, the problem remains unresolved.

\section{Methodology}
\label{method}

Before illustrating more details, we first introduce the definition of feature redundancy as follows:

\begin{definition}
\textbf{Feature redundancy.} A feature is redundant if it is useful for prediction but correlated with another useful feature in the data \cite{guyon2003introduction}.
\end{definition}

\subsection{Problem Statement}
    Given the regularly sampled historical observations $\mathbf{X}=\{\mathbf{x}_t\}_{t=1}^T \in \mathbb{R}^{N\times T}$, where $\mathbf{x}_{t\in T}\in \mathbb{R}^N$ represents $N$ variables at timestamp $t$, our objective is to develop a multi-scale analysis-based method, i.e., DisMS-TS, with the capability to eliminate redundant features of all temporal scales for time series classification. The problem can be formulated as:
	\begin{equation}
		\hat{\mathbf{y}}_{T+1}= \mathcal{G}(\mathbf{X};\Theta_m),
	\end{equation}
	\noindent where $\mathcal{G}(\cdot)$ is the DisMS-TS, which is parameterized by $\Theta_m$, and $\hat{\mathbf{y}}_{T+1}$ is the predicted classification probability at timestamp $t+1$.

    \subsection{Motivation: Feature Redundancy in Multi-scale Time Series}
    \label{motivation}
    As mentioned in Sec.\ref{intro}, due to the significant correlations among features across various temporal scales, the issue of feature redundancy becomes increasingly pronounced \cite{guyon2003introduction}. Let $ \mathcal{F}=\{\mathbf{F}_s\}^{S}_{s=0}$ represents a set of features of $ S+1 $ temporal scales. Due to the existence of redundant scale-shared features in the multi-scale time series, $\mathcal{F}$ can be reformulated as $\mathcal{F}= \{ \mathbf{F}_{sha}+\mathbf{F}_{s,spe}\}_{s=0}^S$, where $\mathbf{F}_{sha}$ denotes the shared feature among all temporal scales, and $\mathbf{F}_{s\in \{0:S\},spe}$ denotes the specific feature of the $s_{th}$ temporal scale. The fundamental assumption in existing multi-scale analysis methods is that future variations are jointly influenced by features from all temporal scales. Ideally, the prediction can be formulated as:
    \begin{equation}
        \mathbf{Y}=\alpha\mathbf{F}_{sha}+\sum_{s=0}^S\alpha_s \mathbf{F}_{s,spe}.
        \label{eq2}
    \end{equation}
    \noindent where $\{ \alpha_s\}_{s=0}^S$ denotes the set of weight coefficients.

    However, due to the entanglement of scale-shared and scale-specific features at individual temporal scales, current methods fail to eliminate these redundant scale-shared features. Essentially, they make predictions by assigning weights to features at each temporal scale through scale-wise fusion methods, such as attention mechanisms. Thus, the prediction process of existing multi-scale methods, with properly allocated weights, can be expressed as:
    
    \begin{equation}
        \begin{aligned}
            \mathbf{Y}&=\sum_{s=0}^S\alpha_s \mathbf{F}_{s}\\
            &=\sum_{s=0}^S \alpha_s(\mathbf{F}_{sha}+\mathbf{F}_{s,spe})\\
            &=(\sum_{s=0}^S \alpha_s)\mathbf{F}_{sha}+\sum_{s=0}^S\alpha_s\mathbf{F}_{s,spe}.
        \end{aligned}
            \label{eq3}
    \end{equation}
    \noindent However, an ideal accurate prediction can be achieved if and only if $\sum_{s=0}^S\alpha_s=\alpha$. When $\sum_{s=0}^S\alpha_s>\alpha$, the model will over-focus on the scale-shared features, while when $\sum_{s=0}^S\alpha_s<\alpha$, the model will under-focus on these scale-shared features.

    Following Eq.\eqref{eq2}, in this study, to avoid the redundancy of scale-shared features, we propose a novel disentanglement approach to learn both scale-shared representation, i.e., $\mathbf{F}_{sha}$, and scale-specific representations, i.e., $\{\mathbf{F}_{s,spe}\}_{s=0}^S$, for enhanced prediction, where $\mathbf{F}_{i,spe}\perp\mathbf{F}_{j,spe}$, for $\forall i, j \in \{0:S\}, i\neq j$.


    \subsection{Overall Framework}
    Figure 2 illustrates the framework of our proposed DisMS-TS. First, we employ a Multi-Scale Generation module (MSG) downsampling the historical observation data into $S$ scales via average pooling. These sampled time series can be viewed as variant series of the original input. After that, the original input, along with the $S$ downsampled time series, is fed into the Multi-scale Temporal Encoder (MTE) to capture temporal dependencies. Subsequently, a Temporal Disentanglement Module (TDM) is proposed to disentangle the temporal representations at each scale, generating both scale-shared and scale-specific temporal representations. Finally, the model is optimized utilizing similarity and disparity losses, combined with classification loss, to enable the predictions based on both scale-shared and scale-specific temporal representations.


	\subsection{Multi-scale Generation Module}
	
	Studies have shown that time series sampled at multiple temporal scales inherently exhibit distinct characteristics, with fine-grained temporal scales containing detailed patterns and coarse-grained scales revealing long-term trend information \cite{mozer1991induction}, which allows us to analyze time series from multiple perspectives.

    Following previous work \cite{wang_2024_Timemixer}, we employ MSG to conduct multi-scale analysis on $\mathbf{X}$, which consists of $S$ average pooling layers, with each layer taking the output of the previous layer as input to generate coarser-grained time series. More specifically, the historical observation $\mathbf{X}$ is fed into MSG, and subsequently, a set of $S+1$ time series $\mathcal{X}=\{\mathbf{X}_s\}_{s=0}^S$ can be obtained, 
    \begin{equation}
    \label{MSG}
        \mathbf{X}_s=AvgPool(\mathbf{X}_{s-1}), s\in\{1,2,...,S\}, 
    \end{equation}
    
    \noindent where $\mathbf{X}_0=\mathbf{X}$ is the original input\footnote{$\mathbf{X}$ and $\mathbf{X}_0$ are defined as equivalent, representing the original input series, and will be used interchangeably in the following content.}, and $\mathbf{X}_{s\in \{0:S\}} \in \mathbb{R}^{\frac{T}{2^s}\times N}$ represents the $s_{th}$-scale time series.  After that, time series at each scale are treated as independent inputs, which are fed into the subsequent modules.
	

    \subsection{Multi-scale Temporal Encoder}
    \label{MTE}
   We employ a simple yet effective multi-scale temporal encoder to capture temporal dependencies in time series across all temporal scales. MTE consists of two sub-modules: a local multi-channel projection (LMP) module for extracting local patterns and a global temporal aggregation (GTA) module for capturing long-range dependencies.
    

    \subsubsection{Local Multi-channel Projection} 

    In contrast to natural language processing, a single time point in a time series typically contains limited semantic information. To address this issue, recent studies partition time series into sub-sequences (i.e., patches) with denser semantic information to capture local temporal dependencies \cite{patchTST}. Furthermore, channel-independence can effectively mitigate the issue of over-fitting. Similar to PatchTST, we utilize a CNN layer to each variable (i.e., channel, as described in other studies) of the time series to capture local patterns, mapping it into a new multi-channel representation. Specifically,

    \begin{equation}	\overline{\mathbf{X}}_{s}^{n}=ReLU(\mathbf{W}\otimes\mathbf{X}_s^n+\mathbf{b}),
    \end{equation}
    
    \noindent where $\mathbf{X}_s^n\in \mathbb{R}^{\frac{T}{2^s}}$ represents the $n_{th}$ variable of the $s_{th}$ temporal scale, $\overline{\mathbf{X}}_{s}^{n}\in \mathbb{R}^{C\times L_s}$ is the local multi-channel representation of the $n_{th}$ variable at the $s_{th}$ temporal scale, $C$ is the number of channels, $L_s$ is the new time steps at the $s_{th}$ scale, $\otimes$ indicates the convolution operator, $\mathbf{W}$ and $\mathbf{b}$ represent the $C$ convolution kernels and biases, respectively.


    

    \subsubsection{Global Temporal Aggregation} 
    Since future temporal variations are influenced by the cumulative effects of historical observations, in this study, we employ the GRU network, which has proven to be effective in capturing temporal dependencies from time series data, to aggregate the historical features of each variable and subsequently generate temporal representation for each scale. Additionally, to account for the varying contributions of each variable, we assign appropriate weights to each variable via an adaptive scoring mechanism. Specifically, $\mathbf{F}_{s\in \{0:S\}}$ can be formulated as:
	
	\begin{equation}
        \begin{aligned}
        {\mathbf{H}}_{s}^{n}&= GRU(\overline{\mathbf{X}}_{s}^{n}),\\
        \mathbf{F}_s&=\underset{n=1}{\overset{N}{\|}} \beta_n \cdot {\mathbf{H}}_{s}^{n}.
        \end{aligned}
        \end{equation}
        
	\noindent where $\|$ indicates concatenation function. $\beta_{n\in\{1: N\}}$is the element of the adaptive learnable vector $\boldsymbol{\beta}$, assigning weights to the temporal features of each variable.

    It is noted that, after downsampling and sub-sequence division, $L_s$ remains small. Thus, we employ GRU instead of Transformer, as the RNN architecture is more effective at capturing temporal dynamics. Moreover, the adaptive scoring mechanism is not required for univariate time series classification tasks.

    \subsection{Temporal Disentanglement Module} 
    
    With the multi-scale temporal representations $\mathbf{F}$, one way is to employ scale-wise feature fusion approaches for aggregation. Although effective, they inherently struggle with the repeated processing of redundant features, leading to suboptimal performance, as discussed in Sec.\ref{motivation}. Thus, we employ a temporal disentanglement approach that decompose $\mathbf{F}_{s\in{\{0:S\}}}$ into scale-shared and scale-specific temporal representations, i.e., $\mathbf{F}_{s,sha}$ and $\mathbf{F}_{s,spe}$. Specifically, 

    
    \begin{equation}
    \label{decompose}
    \begin{aligned}
        \mathbf{M}_{s}&=MLP(\mathbf{F}_s),\\
        \mathbf{M}_{s,sha}&={Sigmoid}(\mathbf{M}_s/\tau)\\
        \mathbf{M}_{s,spe}&={Sigmoid}(-\mathbf{M}_s/\tau)\\
        \mathbf{F}_{s,sha}&=\mathbf{M}_{s,sha}\odot\mathbf{F}_s,
        \mathbf{F}_{s,spe}= \mathbf{M}_{s,spe}\odot\mathbf{F}_s,
        \end{aligned}
     \end{equation} 
     
    \noindent where $MLP(\cdot)$ is a feed forward network with a fully connected layer generating the temporal mask $\mathbf{M}_{s}$. $\mathbf{M}_{s}$ is subsequently processed to generate $\mathbf{M}_{s,sha}$ and $\mathbf{M}_{s,spe}$, which represent the masks of  $\mathbf{F}_{s,sha}$ and $\mathbf{F}_{s,spe}$, respectively. $Sigmoid(\cdot)$ and $\tau$ are the sigmoid function and temperature, respectively. $\mathbf{M}_{s,sha}$ and $\mathbf{M}_{s,spe}$ are each multiplied by $\mathbf{F}_{s}$ to obtain $\mathbf{F}_{s,sha}$ and  $\mathbf{F}_{s,spe}$, respectively, where $\odot$ indicates the dot product. In this way, temporal representations that score higher in scale-shared representations tend to score lower in scale-specific representations, which means $\mathbf{F}_{s,sha}$ and $\mathbf{F}_{s,spe}$ are uncorrelated.

    \subsection{Prediction and Optimization}
    \textbf{Prediction.} Finally, we can utilize scale-shared and scale-specific temporal representations, which contain non-redundant multi-scale features, to make the prediction:
    \begin{equation}
    \label{pred}
        \hat{\mathbf{y}}_{T+1}={Prediction}(\mathbf{F}_{sha}||\mathbf{F}_{spe}),
    \end{equation}

    \noindent where $\mathbf{F}_{sha}=\frac{1}{S+1}\sum_{s=0}^S\mathbf{F}_{s,sha}$, $\mathbf{F}_{spe}= \underset{s=0}{\overset{S}{\|}} \mathbf{F}_{s,spe}$, $\|$ indicates the concatenation operator, and $Prediction(\cdot)$ is a feed forward network with two fully-connected layers.

   \noindent\textbf{Optimization.}
   To empower the DisMS-TS with the capability to effectively learn both scale-shared and scale-specific temporal representations, we introduce two regularization terms to constrain their feature distributions. First, we propose a similarity regularization term to ensure that all scale-shared temporal representations exhibit a consistent distribution, i.e., $\forall i,j\in \{0:S\}, \mathbf{F}_{i,sha}=\mathbf{F}_{j,sha}$. Specifically, 

        \begin{equation}
        \label{sim}
    \small
        \mathcal{L}_{sim}=\frac{2}{S{(S+1)}} \sum_{i=0}^{S-1}\sum_{j=i+1}^S {MSE}({Sim}(\mathbf{F}_{i,sha},\mathbf{F}_{j,sha}),1),
    \end{equation}

    \noindent where $MSE(\cdot)$ denotes the mean squared error function and $Sim(\cdot)$ refers to the cosine similarity. 
    
    Second, the disparity regularization term is designed to ensure that all scale-specific representations are orthogonal and maintain independent distributions, i.e., $\forall i, j \in \{0:S\}, i\neq j$, $\mathbf{F}_{i,spe}\perp\mathbf{F}_{j,spe}$, which is formulated as:
    \begin{equation}
    \label{dis}
    \small
        \mathcal{L}_{dis}=\frac{2}{S{(S+1)}} \sum_{i=0}^{S-1}\sum_{j=i+1}^S {MSE}({Sim}(\mathbf{F}_{i,spe},\mathbf{F}_{j,spe}),0).
    \end{equation}

    \noindent The final training objective is,
    \begin{equation}
    \label{loss}
        \small
        \mathcal{L}=\mathcal{L}_{class}+\lambda_1\mathcal{L}_{sim}+\lambda_2\mathcal{L}_{dis}
    \end{equation}
    \noindent where $\mathcal{L}_{class}$ is the cross-entropy loss, $\lambda_1$ and $\lambda_2$ are the hyperparameters striking a balance between three terms. 
    
    The algorithm of  DisMS-TS is shown in Algorithm 1.

\begin{algorithm}[t]
    \label{algorithm1}
\caption{Training pipeline for DisMS-TS}
\begin{algorithmic}[1]  
\State \textbf{Input:} Time series $\mathbf{X}$, training epochs $L$, scale number $S$, hyperparameter $\lambda_1$, $\lambda_2$.
\State \textbf{Output:} Optimized parameters $\Theta_m$, prediction of $\hat{\mathbf{y}}_{T+1}$.
\State Randomly initialize the model parameters.
\For{$l=1,2,...,L$}
    \State Obtain multi-scale series $\{\mathbf{X}_s\}_{s=0}^S$ as Eq.\eqref{MSG}.
    \State Generate temporal representations $\{\mathbf{F}_s\}_{s=0}^S$ as Sec.\ref{MTE}.
    \State Obtain $\{\mathbf{F}_{s,sha}\}_{s=0}^S$ and $\{\mathbf{F}_{s,spe}\}_{s=0}^S$ as Eq.\eqref{decompose}.
    \State Make prediction as Eq.\eqref{pred}.
    \State Calculate similarity and disparity loss as Eq.\eqref{sim} \& Eq.\eqref{dis}.
    \State Update the model according to Eq.\eqref{loss}.
\EndFor
\end{algorithmic}
\end{algorithm}





	

\section{Experiments}
In this section, we conduct experiments for our proposed DisMS-TS, aiming to address the following research questions:
\begin{itemize}
    \item \textbf{RQ1:} How is the overall performance of DisMS-TS as compared to various classic and state-of-the-art methods?
    \item \textbf{RQ2:} How does each component of DisMS-TS contribute to the performance?
    \item \textbf{RQ3:} How do hyperparameter confgurations affect the performance of DisMS-TS?
    \item \textbf{RQ4:} Does DisMS-TS effectively disentangle $\mathbf{F}$ into $\mathbf{F}_{sha}$ and $\mathbf{F}_{spe}$ through the proposed disentanglement module and regularization terms?
\end{itemize}

\subsection{Experimental Setup} 

\subsubsection{Datasets} To evaluate the effectiveness of our proposed method, we conduct extensive experiments on six time series classification benchmark datasets, encompassing diverse application domains. For multivariate time series classification, we utilize five datasets, including Human Activity Recognition (HAR\footnote{https://paperswithcode.com/dataset/har}), ISRUC-S3\footnote{https://sleeptight.isr.uc.pt/}, NASDAQ\footnote{https://www.nasdaq.com/}, and two datasets from the UEA archive\footnote{https://www.timeseriesclassification.com/}, i.e., Articulary Word Recognition (AWR) and the Finger Movements (FM). We also conduct experiments on a univariate time series classification dataset, Epileptic Seizure Recognition dataset (Epilepsy\footnote{https://www.kaggle.com/datasets/harunshimanto/epileptic-seizure-recognition}). The training, validation, and test sets for HAR, ISRUC-S3, NASDAQ, and Epilepsy are divided following the approach employed in related works \cite{eldele2021time,wang2024graph}. For datasets from UEA archive, we employ their predefined train-test splits. Table \ref{dataset} summarizes the details of each dataset.

\begin{table}[t]
    \centering
\renewcommand{\arraystretch}{0.85}
    \begin{tabular}{c|ccccc}
    \toprule[1.pt]
    Dataset&\#Train&\#Test&Length&\#Variable&\#Class\\
        \toprule[.7pt]
         HAR&5881&2947&128&9&6\\
         ISRUC-S3&6013&1718&3000&10&5\\ 
         NASDAQ&2387&682&128&5&3\\
         AWR&275&300&144&9&25\\
         FM&316&100&50&10&2\\
         Epilepsy&7360&2300&178&1&5\\
    \bottomrule[1.pt]
    \end{tabular}
    \caption{Dataset statistics.}
\label{dataset}
\end{table}

\subsubsection{Evaluation Metrics} In alignment with related researches \cite{eldele2021time}, we adopt accuracy (ACC) and macro-averaged F1-score (F1) as primary evaluation metrics. Furthermore, we incorporate the Matthews correlation coefficient (MCC), which proves particularly valuable in situations where the class distributions are imbalanced. Higher values of these metrics indicate superior performance.

\subsubsection{Baseline Methods} To validate the effectiveness of the proposed DisMS-TS, we evaluate it against seven well-known methods, including conventional non-multi-scale analysis-based methods such as LSTNet \cite{lai2018lstnet}, DTP \cite{lee2021DPT}, TS-TCC \cite{eldele2021time} and PatchTST \cite{patchTST}, as well as the latest baselines based on multi-scale time series analysis, including MAGNN \cite{chen_2023_MAGNN}, Pathformer \cite{chen2024pathformer}, and Timemixer \cite{wang_2024_Timemixer}. All methods are introduced in Sec.\ref{relaredwork} and re-implemented according to their original configurations.

\begin{table*}[t]
    \renewcommand{\arraystretch}{0.7}
    \centering
    \begin{tabular}
        {c|c|cccccc}
        \toprule[1.pt]
        Methods& Metrics& HAR&ISRUC-S3&NASDAQ&AWR&FM&Epilepsy\\ 
        \toprule[1.pt]
        
        \multirow{3}{*}{LSTNet [2018]}&
         ACC(\%)&87.52$\pm$1.25&80.19$\pm$1.21&35.34$\pm$1.60&91.73$\pm$0.88&51.60$\pm$3.01&62.57$\pm$1.08\\
        &F1(\%)&86.58$\pm$1.28&77.93$\pm$1.12&25.29$\pm$2.18&91.33$\pm$0.93&51.59$\pm$3.00&60.36$\pm$1.72\\
        &MCC&0.846$\pm$0.014&0.746$\pm$0.016&0.058$\pm$0.022&0.914$\pm$0.009&0.032$\pm$0.060&0.556$\pm$0.011\\
        \midrule
        \multirow{3}{*}{DTP [2021]}&
         ACC(\%)&92.09$\pm$0.88&74.23$\pm$1.04&34.25$\pm$0.85&94.87$\pm$1.20&50.00$\pm$1.98&42.57$\pm$2.46\\
        &F1(\%)&92.10$\pm$0.85&72.38$\pm$1.05&22.78$\pm$1.14&94.82$\pm$1.23&49.98$\pm$1.99&41.14$\pm$2.17\\
        &MCC&0.905$\pm$0.010&0.671$\pm$0.012&0.046$\pm$0.015&0.946$\pm$0.012&0.020$\pm$0.039&0.301$\pm$0.036\\
        
        \midrule
        \multirow{3}{*}{TS-TCC [2021]}&
         ACC(\%)&90.62$\pm$0.59&77.63$\pm$1.18&35.52$\pm$1.26&89.65$\pm$0.58&51.60$\pm$1.62&64.84$\pm$1.14\\
        &F1(\%)&90.66$\pm$0.62&75.50$\pm$1.17&24.96$\pm$1.52&89.57$\pm$0.60&51.33$\pm$1.80&63.26$\pm$1.42\\
        &MCC&0.898$\pm$0.006&0.735$\pm$0.016&0.054$\pm$0.017&0.889$\pm$0.007&0.030$\pm$0.032&0.552$\pm$0.015\\

        \midrule
        \multirow{3}{*}{PatchTST [2023]}&
         ACC(\%)&93.34$\pm$0.32&81.05$\pm$0.81&36.51$\pm$1.06&97.47$\pm$0.57&53.40$\pm$3.61&68.68$\pm$1.56\\
        &F1(\%)&93.28$\pm$0.34&79.14$\pm$0.80&25.50$\pm$1.27&97.45$\pm$0.64&52.83$\pm$3.68&67.26$\pm$1.72\\
        &MCC&0.919$\pm$0.003&0.767$\pm$0.015&0.056$\pm$0.016&0.965$\pm$0.005&0.106$\pm$0.071&0.609$\pm$0.020\\
        
        \midrule
        \multirow{3}{*}{MAGNN [2023]}&
         ACC(\%)&92.45$\pm$0.58&79.07$\pm$0.86&38.12$\pm$1.13&93.33$\pm$0.61&52.20$\pm$2.48&51.42$\pm$2.20\\
        &F1(\%)&92.52$\pm$0.64&77.51$\pm$0.85&29.52$\pm$1.46&93.36$\pm$0.75&51.32$\pm$1.74&49.55$\pm$2.69\\
        &MCC&0.912$\pm$0.008&0.749$\pm$0.012&0.079$\pm$0.014&0.930$\pm$0.006&0.044$\pm$0.055&0.430$\pm$0.027\\

        \midrule
        \multirow{3}{*}{Pathformer [2024]}&
         ACC(\%)&94.88$\pm$0.76&82.52$\pm$0.78&37.24$\pm$0.89&\textbf{98.43$\pm$1.13}&\underline{55.60$\pm$1.85}&\underline{69.18$\pm$0.88}\\
        &F1(\%)&94.92$\pm$0.74&81.27$\pm$0.83&27.96$\pm$1.62&\textbf{98.41$\pm$1.15}&\underline{54.92$\pm$1.90}&\underline{67.28$\pm$0.85}\\
        &MCC&0.938$\pm$0.008&0.771$\pm$0.012&0.078$\pm$0.016&\textbf{0.981$\pm$0.013}&\underline{0.121$\pm$0.041}&\underline{0.614$\pm$0.010}\\

        \midrule
        \multirow{3}{*}{Timemixer [2024]}&
        ACC(\%)&\underline{95.53$\pm$0.65}&\underline{83.50$\pm$1.04}&\underline{38.42$\pm$0.75}&97.53$\pm$0.98&54.20$\pm$3.73&67.28$\pm$1.19\\
        &F1(\%)&\underline{95.60$\pm$0.64}&\underline{82.18$\pm$1.12}&\underline{29.68$\pm$0.97}&97.54$\pm$1.01&53.67$\pm$3.85&65.65$\pm$1.35\\
        &MCC&\underline{0.947$\pm$0.005}&\underline{0.790$\pm$0.014}&\underline{0.088$\pm$0.012}&0.972$\pm$0.010&0.082$\pm$0.074&0.584$\pm$0.015\\

        \midrule
        \multirow{3}{*}{\shortstack{ DisMS-TS\\(Our proposed)}}&
        ACC(\%)&\textbf{96.47$\pm$0.37}&\textbf{85.37$\pm$0.51}&\textbf{39.71$\pm$0.76}&\underline{98.27$\pm$0.86}&\textbf{61.00$\pm$3.16}&\textbf{71.30$\pm$0.71}\\
         &F1(\%)&\textbf{96.43$\pm$0.40}&\textbf{84.20$\pm$0.54}&\textbf{32.69$\pm$0.833}&\underline{98.23$\pm$0.90}&\textbf{60.67$\pm$3.10}&\textbf{70.70$\pm$1.18}\\
         &MCC&\textbf{0.957$\pm$0.004}&\textbf{0.829$\pm$0.008}&\textbf{0.104$\pm$0.012}&\underline{0.976$\pm$0.009}&\textbf{0.228$\pm$0.065}&\textbf{0.644$\pm$0.006}\\

        \midrule
        \multirow{3}{*}{Improvements}
        &ACC(\%)&$\uparrow$ 0.98\%*&$\uparrow$ 2.24\%*&$\uparrow$ 3.35\%*&$\downarrow$ 0.16\%&$\uparrow$ 9.71\%*&$\uparrow$ 3.06\%*\\
        &F1(\%)&$\uparrow$ 0.86\%*&$\uparrow$ 2.45\%*&$\uparrow$ 10.14\%*&$\downarrow$ 0.18\%&$\uparrow$ 10.47\%*&$\uparrow$ 5.08\%*\\
         &MCC&$\uparrow$ 1.05\%*&$\uparrow$ 4.93\%*&$\uparrow$ 18.18\%*&$\downarrow$ 0.50\%&$\uparrow$ 88.4\%*&$\uparrow$ 4.78\%*\\
        \bottomrule[1pt]
        
    \end{tabular}
    \label{Resultcomparison}
    \caption{Performance comparison with baselines on six datasets. Bold \& underline show the best \& second best results, respectively. * indicates the improvement to the best baseline is statistically significant (t-test with p-value <0.01).}
\end{table*}

\subsubsection{Implementation Details}
All experiments are conducted on a single NVIDIA GeForce RTX 4070Ti GPU using PyCharm 3.8 and PyTorch 2.1. The batch size is set to 256, except for the ISRUC-S3 dataset, where it is set to 48 due to the larger size of the dataset. The ADAM optimizer \cite{kingma2014adam} is applied with an initial learning rate of 5e-3, and each model undergoes training for 100 epochs. Additionally, each experiment is conducted independently five times, with the mean and standard deviation reported for comparison.

\subsection{Results and Analysis (RQ1)}
Table 2 illustrates the comparison with baseline methods, from which we can easily observe that our method achieves the optimal performance overall. Specifically, to validate the reliability of the proposed DisMS-TS, we independently conducted five experiments and performed a t-test at a significance level of $\alpha=0.01$ to assess the performance differences between DisMS-TS and other baseline methods. The results demonstrate that DisMS-TS significantly outperforms ($p<0.01$) its counterparts across five datasets, except for the AWR dataset, where it exhibits suboptimal performance compared to Pathformer. By decoupling the multi-scale time series into multi-scale seasonal and trend components to achieve multi-scale mixing, Timemixer obtains suboptimal performance on HAR, ISRUC-S3, and NASDAQ datasets. Pathformer, which captures global correlations and local details in time series through adaptive pathways and multi-scale division, achieves suboptimal performance on datasets FM and Epilepsy datasets. On the HAR, ISRUC-S3, and NASDAQ datasets, DisMS-TS demonstrates significant improvements, achieving increases of 0.98\%, 2.24\%, and 3.35\% in ACC, respectively, compared to the suboptimal Timemixer. Moreover, DisMS-TS achieves improvements over the suboptimal Pathformer, with increases of 9.71\% and 3.06\% on the FM and Epilepsy datasets, respectively. On the AWR dataset, the ACC, F1, and MCC of our method are slightly 0.16\%, 0.18\%, and 0.5\% lower than the optimal Pathformer. Additionally, it was also observed that in the univariate time series prediction scenario, MAGNN achieves poor performance on the Epilepsy dataset due to its over-reliance on the relationships between variables.

\begin{table*}[h]
    \renewcommand{\arraystretch}{0.7}
    \centering
    \begin{tabular}
        {c|c|cccccc}
        \toprule[1.pt]
        Methods& Metrics& HAR&ISRUC-S3&NASDAQ&AWR&FM&Epilepsy\\ 
        \toprule[1.pt]
        \multirow{3}{*}{Without LMP}
        &ACC(\%)&90.29$\pm$0.14&75.67$\pm$1.55&35.29$\pm$0.90&93.84$\pm$0.68&51.40$\pm$1.83&61.43$\pm$0.86\\
        &F1(\%)&90.43$\pm$0.12&72.49$\pm$1.42&22.35$\pm$1.43&92.57$\pm$0.45&51.06$\pm$1.64&59.77$\pm$0.84\\
        &MCC&0.884$\pm$0.001&0.662$\pm$0.026&0.053$\pm$0.021&0.920$\pm$0.007&0.032$\pm$0.032&0.552$\pm$0.012\\

        \midrule
        \multirow{3}{*}{\shortstack{Replaced with\\ MAGNN's SWF}}&ACC(\%)&91.85$\pm$1.76&81.92$\pm$0.95&38.46$\pm$1.17&97.39$\pm$0.89&52.50$\pm$2.59&67.33$\pm$0.47\\
         &F1(\%)&91.82$\pm$1.76&81.22$\pm$1.08&28.68$\pm$1.49&97.17$\pm$0.81&51.87$\pm$2.23&66.60$\pm$0.91\\
         &MCC&0.902$\pm$0.020&0.778$\pm$0.14&0.083$\pm$0.015&0.965$\pm$0.011&0.062$\pm$0.057&0.603$\pm$0.005\\
         \midrule
        \multirow{3}{*}{\shortstack{Replaced with\\ Timemixer's SWF}}&
        ACC(\%)&95.10$\pm$1.24&83.81$\pm$0.78&38.56$\pm$0.91&97.96$\pm$1.07&56.60$\pm$3.27&68.06$\pm$0.96\\
         &F1(\%)&95.06$\pm$1.26&82.72$\pm$0.69&28.69$\pm$0.80&97.74$\pm$0.96&55.75$\pm$2.85&68.69$\pm$0.96\\
         &MCC&0.941$\pm$0.014&0.809$\pm$0.012&0.088$\pm$0.014&0.967$\pm$0.013&0.175$\pm$0.062&0.620$\pm$0.014\\
         \midrule

        \multirow{3}{*}{DisMS-TS}&
        ACC(\%)&\textbf{96.47$\pm$0.37}&\textbf{85.37$\pm$0.51}&\textbf{39.71$\pm$0.76}&\textbf{98.27$\pm$0.86}&\textbf{61.00$\pm$3.16}&\textbf{71.30$\pm$0.71}\\
         &F1(\%)&\textbf{96.43$\pm$0.40}&\textbf{84.20$\pm$0.54}&\textbf{32.69$\pm$0.833}&\textbf{98.23$\pm$0.90}&\textbf{60.67$\pm$3.10}&\textbf{70.70$\pm$1.18}\\
         &MCC&\textbf{0.957$\pm$0.004}&\textbf{0.829$\pm$0.008}&\textbf{0.104$\pm$0.012}&\textbf{0.976$\pm$0.009}&\textbf{0.228$\pm$0.065}&\textbf{0.644$\pm$0.006}\\
	    \bottomrule[1pt]
        
    \end{tabular}
    \label{11}
    \caption{Ablation study.}
\end{table*}



Furthermore, methods relying on multi-scale analysis outperform those based on non-multi-scale analysis in terms of overall performance. This can be attributed to the fact that multi-scale analysis provides more temporal patterns, such as coarse-grained trend information. This is particularly evident in the stock trend prediction scenario on the NASDAQ dataset, where investors typically forecast future stock trends relying on daily, weekly, and monthly technical indicators. Nevertheless, existing multi-scale analysis-based time series predictors either directly utilize feature fusion approaches to integrate features from all temporal scales or select optimal $K$-scale features for prediction, both of which fail to eliminate redundant features among multi-scale time series. For the latter approaches, when $K=1$, although no redundant features exist, the temporal patterns from other scales are frequently ignored. In summary, our proposed DisMS-TS not only focuses on eliminating redundant scale-shared features across multi-scale time series but also considers temporal features from multi-scale perspectives.

\subsection{Ablation Study (RQ2)}
We conduct an ablation study to assess the effectiveness of each component of DisMS-TS. In the first variant, we exclude the usage of LMP module from DisMS-TS. Moreover, to verify the effectiveness of disentangling multi-scale features into $\mathbf{F}_{sha}$ and $\mathbf{F}_{spe}$ for prediction, we replace the temporal disentanglement module and the proposed regularization terms with scale-wise fusion (SWF) approaches from existing methods (i.e., MAGNN \cite{chen_2023_MAGNN} and TimeMixer \cite{wang_2024_Timemixer}), respectively, while keeping the multi-scale temporal encoder unchanged.

Table 3 demonstrates the ablation study across six datasets. Obviously, DisMS-TS achieves the superior performance, which showcases the positive impact of each module on prediction performance. Specifically, we take the ACC results as examples. Comparing against the $``$Without LMP’’ variant, we observe that our proposed DisMS-TS achieves an improvement ranging from 6.84\% to 18.67\% across six datasets, which verifies the importance of the LMP module in capturing the local patterns of sub-sequences for each variable. With the introduction of the TDM and the regularization terms, which explicitly disentangle $\mathbf{F}$ into $\mathbf{F}_{sha}$ and $\mathbf{F}_{spe}$, a noticeable performance boost is observed compared to directly using SWF approaches. Specifically, it achieves the accuracy improvements of up to 16.19\% and 7.77\%, respectively, compared to the $``$Replaced with MAGNN’s SWF’’ and $``$Replaced with Timemixer’s SWF’’ variants across six datasets. This validates the previous analysis in Sec.\ref{motivation}, that eliminating redundant shared features across scales is essential for multi-scale analysis-based methods.

\subsection{Hyperparameter Analysis (RQ3)}

\subsubsection{The Number of Temporal Scales $S$}
To assess the impact of $S$ on prediction performance, we conduct experiments on four datasets, i.e., HAR, ISRUC-S3, NASDAQ and Epilepsy with various configurations of $S \in \{0, 1, 2, 3, 4, 5, 6, 7\}$. It is noted that $S=0$ indicates the absence of multi-scale analysis. Figure \ref{all_S} illustrates the experimental results, from which we obtain several observations. First, multi-scale analysis plays a significant role in enhancing the model's predictive performance. Second, the performance of different $S$ varies across different datasets. Specifically, when $S$ equals 3, 5, 5, and 6 the optimal results are achieved on the HAR, ISRUC-S3, NASDAQ and Epilepsy datasets, respectively. When handling time series with longer time steps, a greater number of downsampling levels is required. When the prediction task is trend-related, such as stock trend classification, utilizing more downsampling layers may better capture the future trend characteristics of the time series. However, a very large $S$ may result in the extraction of excessive coarse-grained features, which hinders the model's ability to capture scale-shared features and consequently impairing prediction performance.

\subsubsection{The Value of $\lambda_1$ and $\lambda_2$ (RQ4)} We further discuss the sensitivity of the hyperparameters $\lambda_1$ and $\lambda_2$ in Eq.\ref{loss}, which balance the trade-offs between classification, similarity and disparity losses. In this study, we set $\lambda_1$ and $\lambda_2$ to be equal, and both are uniformly denoted as $\lambda$. We choose the values within \{0.001, 0.005, 0.01, 0.05, 0.1, 0.5, 1.0\}, and test the performance on HAR and Epilepsy datasets. We can observe from Figure \ref{all_lambda} that DisMS-TS tends to achieve better performance when $\lambda$ is set to higher values, which demonstrates that the similarity and disparity losses both have a positive impact on prediction performance. However, as the values increase beyond a certain threshold, the improvements become less pronounced.


\begin{figure}[t]
    \centering
    \begin{subfigure}[b]{0.23\textwidth}
        \centering
        \includegraphics[width=\linewidth]{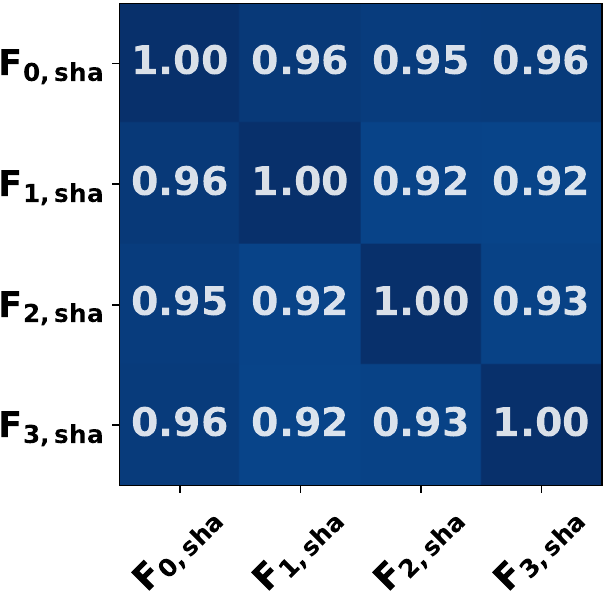}
        \caption{HAR}
    \end{subfigure} 
    \hspace{-0.005\textwidth}
    \begin{subfigure}[b]{0.23\textwidth}
        \centering
        \includegraphics[width=\linewidth]{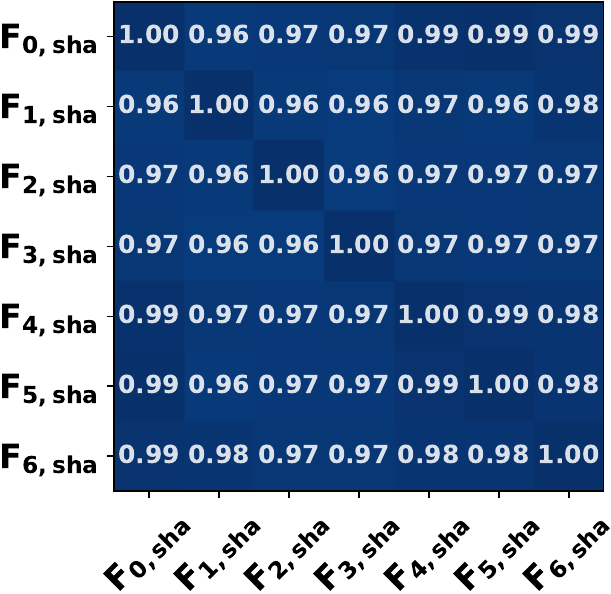}
        \caption{Epilepsy}
    \end{subfigure}

    \caption{Correlation analysis of scale-shared representations.} 
    \label{Shared_analysis_matrix}

\end{figure}

\begin{figure}[t]
    \centering
    \begin{subfigure}[b]{0.23\textwidth}
        \centering
        \includegraphics[width=\linewidth]{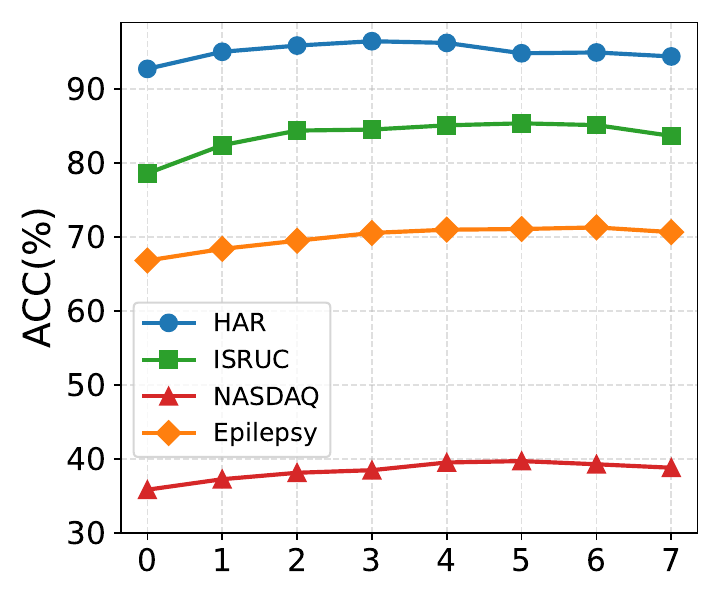}
        \caption{Effect of different $S$.}
    \label{all_S}
    \end{subfigure} 
    \begin{subfigure}[b]{0.23\textwidth}
        \centering
        \includegraphics[width=\linewidth]{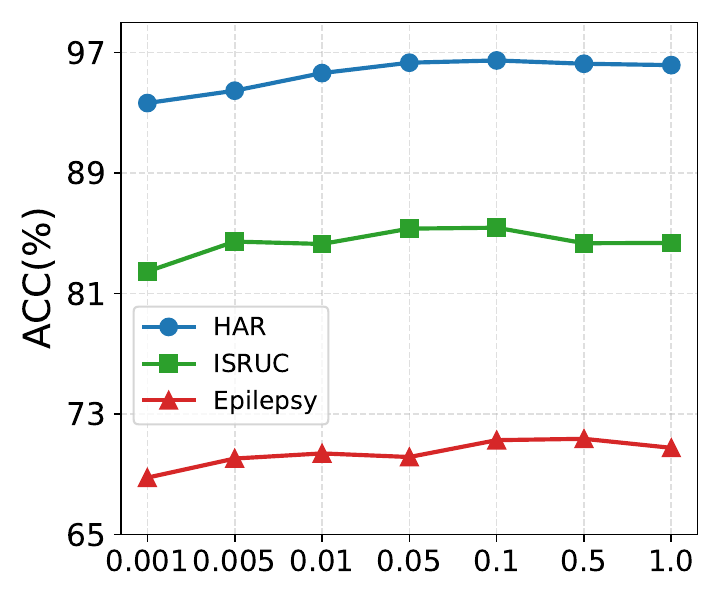}
        \caption{Effect of different $\lambda$.}
    \label{all_lambda}
    \end{subfigure}

    \caption{Hyperparameter analysis.} 
    
    \label{lambda}
\end{figure}

\subsection{Analysis of $\mathbf{F}_{sha}$ and $\mathbf{F}_{spe}$}

\subsubsection{Correlation Analysis.} 
\begin{figure}[t]
    \centering
    \begin{subfigure}[b]{0.23\textwidth}
        \centering
        \includegraphics[width=\linewidth]{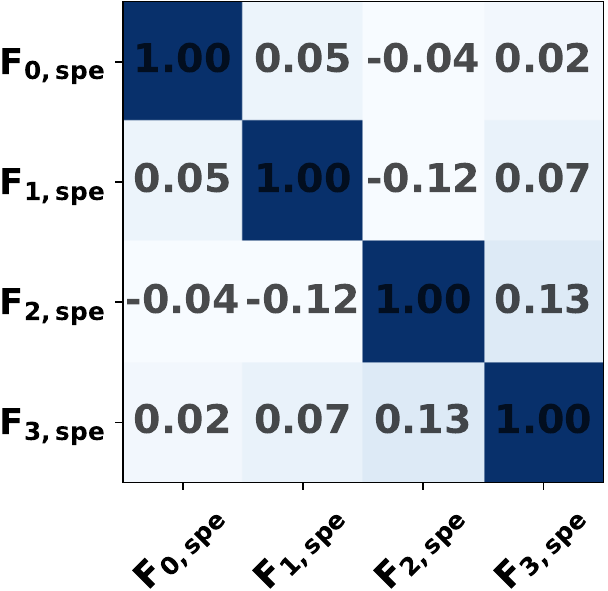}
        \caption{HAR}
    \end{subfigure} 
    \hspace{-0.005\textwidth}
    \begin{subfigure}[b]{0.23\textwidth}
        \centering
        \includegraphics[width=\linewidth]{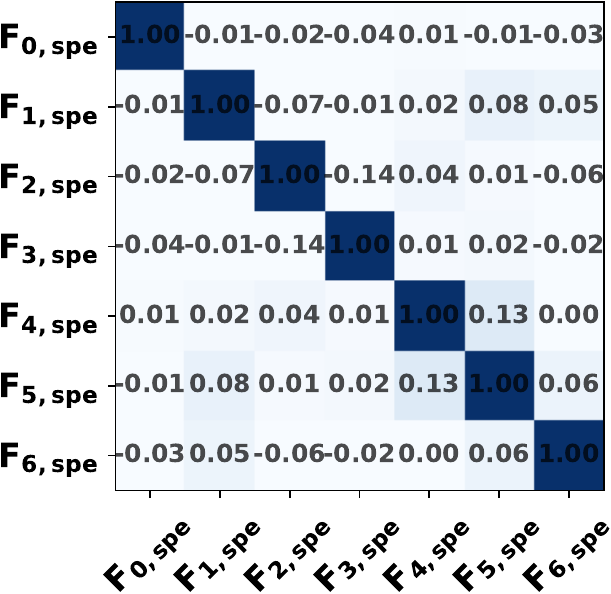}
        \caption{Epilepsy}
    \end{subfigure}

    \caption{Correlation analysis of scale-specific representations.} 
    \label{Specific_analysis}
\end{figure}

To further verify whether DisMS-TS can effectively disentangle $\mathbf{F}$ into $\mathbf{F}_{sha}$ and $\mathbf{F}_{spe}$, we conduct correlation analysis on them, respectively. Figure \ref{Shared_analysis_matrix} and Figure \ref{Specific_analysis} illustrate the corresponding correlation matrices on the multivariate HAR and univariate Epilepsy datasets. The elements in the matrix represent the values of cosine similarity between temporal representations. From Figure \ref{Shared_analysis_matrix}, we can observe that for any pair of scale-shared representations $\mathbf{F}_{i\in {\{0:S\}},sha}$ and $\mathbf{F}_{j\in {\{0:S\}},sha}$ ($i\neq j$), they exhibit an extremely  high correlation. Moreover, we can also observe that for any pair of scale-specific representations $\mathbf{F}_{i\in {\{0:S\}},spe}$ and $\mathbf{F}_{j\in {\{0:S\}},spe}$ ($i\neq j$), they are highly uncorrelated.

\subsubsection{Visualization Analysis.} 
\begin{figure}[t]
    \centering
    \begin{minipage}{0.23\textwidth}
        \centering
        \includegraphics[width=\textwidth]{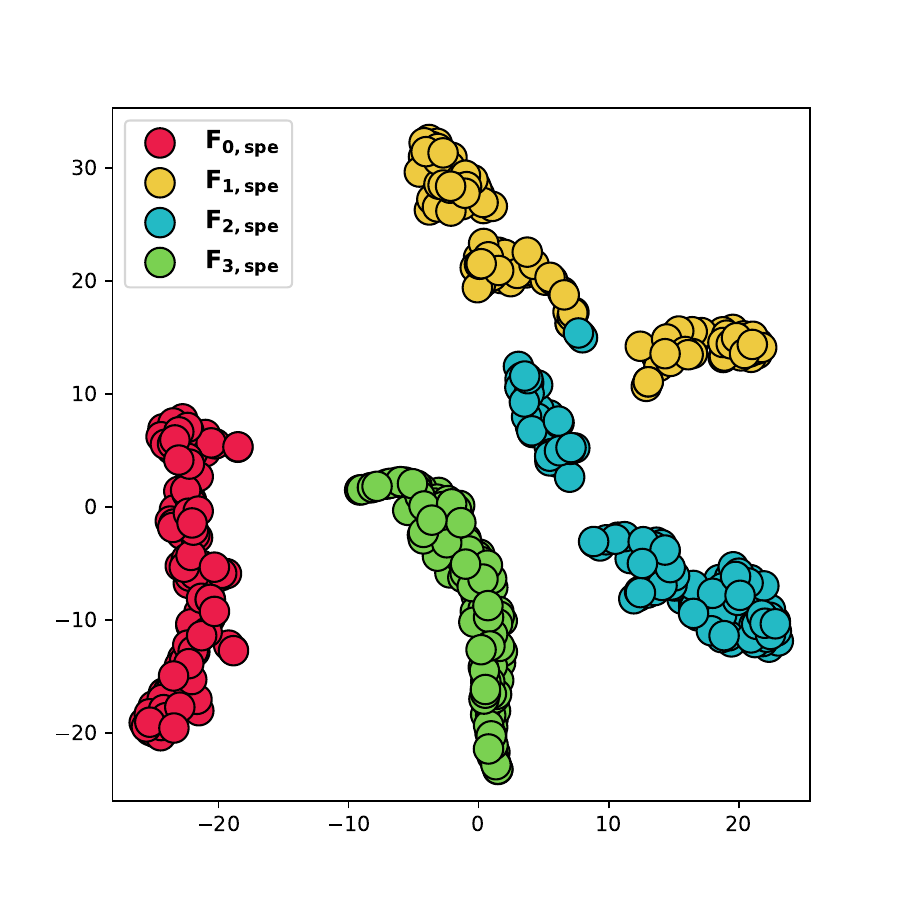}
        \subcaption{HAR}\label{fig:sub1}
    \end{minipage}\hspace{0.005\textwidth}
    \begin{minipage}{0.23\textwidth}
        \centering
        \includegraphics[width=\textwidth]{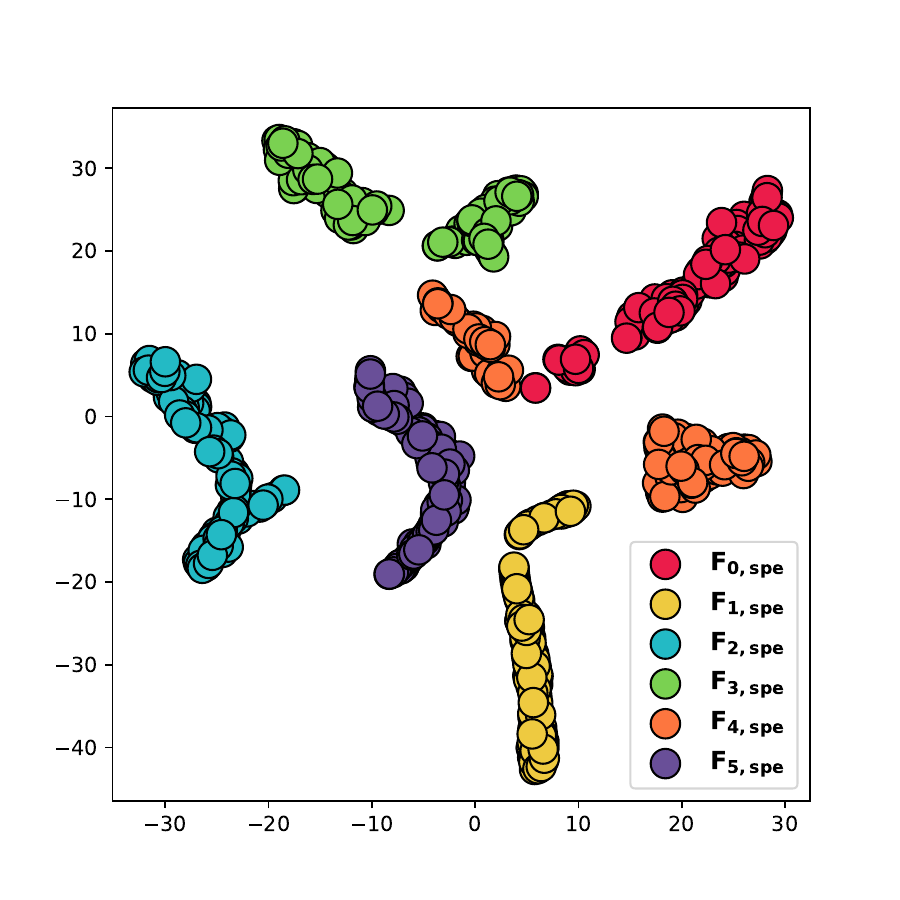}
        \subcaption{Epilepsy}\label{fig:sub2}
    \end{minipage}
    \caption{The t-SNE Visualization of $\mathbf{F}_{spe}$ on two datasets.}
    \label{visilization}
\end{figure}

To explicitly assess the quality of $\mathbf{F}_{spe}$ learned by DisMS-TS, we make a visualization by projecting $\mathbf{F}_{spe}$ into a two-dimensional space using t-SNE, as shown in Figure \ref{visilization}. Specifically, we conduct t-SNE analysis on scale-specific temporal representations at four and six scales for the HAR and Epilepsy datasets, respectively. We can observe that scale-specific temporal features from different scales tend to be clustered into distinct regions, with minimal overlap between them. Overall, the results demonstrate that DisMS-TS can eliminate redundant shared features in multi-scale time series to a great extent.

\section{Conclusion}
In this paper, we focused on the performance issues of multi-scale analysis-based methods in the task of time series classification. We found and analyzed that conventional multi-scale analysis-based methods fail to eliminate redundant multi-scale features, which can jeopardize prediction performance. To address this issue, we proposed a novel disentangled multi-scale framework, called DisMS-TS, for time series classification. Specifically,  we propose a temporal disentanglement module that disentangles the multi-scale temporal representation into scale-shared and scale-specific representations for prediction, and two regularization terms (i.e., similarity loss and disparity loss) to ensure the consistency of scale-shared representations and the disparity of scale-specific representations across all temporal scales, respectively. Through extensive experiments on multiple datasets, we demonstrated the effectiveness and superiority of the proposed DisMS-TS.

\section*{Acknowledgments}
This study is supported by Northeastern University, Shenyang, China (02110022124005).

    
    
    

\bibliographystyle{ACM-Reference-Format}
\bibliography{sample-base}


\begin{thebibliography}{57}


\ifx \showCODEN    \undefined \def \showCODEN     #1{\unskip}     \fi
\ifx \showISBNx    \undefined \def \showISBNx     #1{\unskip}     \fi
\ifx \showISBNxiii \undefined \def \showISBNxiii  #1{\unskip}     \fi
\ifx \showISSN     \undefined \def \showISSN      #1{\unskip}     \fi
\ifx \showLCCN     \undefined \def \showLCCN      #1{\unskip}     \fi
\ifx \shownote     \undefined \def \shownote      #1{#1}          \fi
\ifx \showarticletitle \undefined \def \showarticletitle #1{#1}   \fi
\ifx \showURL      \undefined \def \showURL       {\relax}        \fi
\providecommand\bibfield[2]{#2}
\providecommand\bibinfo[2]{#2}
\providecommand\natexlab[1]{#1}
\providecommand\showeprint[2][]{arXiv:#2}

\bibitem[Chen et~al\mbox{.}(2023)]%
        {chen_2023_MAGNN}
\bibfield{author}{\bibinfo{person}{Ling Chen}, \bibinfo{person}{Donghui Chen},
  \bibinfo{person}{Zongjiang Shang}, \bibinfo{person}{Binqing Wu},
  \bibinfo{person}{Cen Zheng}, \bibinfo{person}{Bo Wen}, {and}
  \bibinfo{person}{Wei Zhang}.} \bibinfo{year}{2023}\natexlab{}.
\newblock \showarticletitle{Multi-scale adaptive graph neural network for
  multivariate time series forecasting}.
\newblock \bibinfo{journal}{\emph{IEEE Transactions on Knowledge and Data
  Engineering}} \bibinfo{volume}{35}, \bibinfo{number}{10}
  (\bibinfo{year}{2023}), \bibinfo{pages}{10748--10761}.
\newblock


\bibitem[Chen et~al\mbox{.}(2024)]%
        {chen2024pathformer}
\bibfield{author}{\bibinfo{person}{Peng Chen}, \bibinfo{person}{Yingying
  Zhang}, \bibinfo{person}{Yunyao Cheng}, \bibinfo{person}{Yang Shu},
  \bibinfo{person}{Yihang Wang}, \bibinfo{person}{Qingsong Wen},
  \bibinfo{person}{Bin Yang}, {and} \bibinfo{person}{Chenjuan Guo}.}
  \bibinfo{year}{2024}\natexlab{}.
\newblock \showarticletitle{Pathformer: Multi-scale transformers with adaptive
  pathways for time series forecasting}.
\newblock \bibinfo{journal}{\emph{arXiv preprint arXiv:2402.05956}}
  (\bibinfo{year}{2024}).
\newblock


\bibitem[Chen and Shi(2021)]%
        {chen_2021_multiscale_Neural_Networks}
\bibfield{author}{\bibinfo{person}{Wei Chen} {and} \bibinfo{person}{Ke Shi}.}
  \bibinfo{year}{2021}\natexlab{}.
\newblock \showarticletitle{Multi-scale attention convolutional neural network
  for time series classification}.
\newblock \bibinfo{journal}{\emph{Neural Networks}}  \bibinfo{volume}{136}
  (\bibinfo{year}{2021}), \bibinfo{pages}{126--140}.
\newblock


\bibitem[Chen et~al\mbox{.}(2021)]%
        {chen2021scaleselect1}
\bibfield{author}{\bibinfo{person}{Zipeng Chen}, \bibinfo{person}{Qianli Ma},
  {and} \bibinfo{person}{Zhenxi Lin}.} \bibinfo{year}{2021}\natexlab{}.
\newblock \showarticletitle{Time-Aware Multi-Scale RNNs for Time Series
  Modeling.}. In \bibinfo{booktitle}{\emph{IJCAI}}.
  \bibinfo{pages}{2285--2291}.
\newblock


\bibitem[Cheng et~al\mbox{.}(2023b)]%
        {cheng2023www}
\bibfield{author}{\bibinfo{person}{Mingyue Cheng}, \bibinfo{person}{Qi Liu},
  \bibinfo{person}{Zhiding Liu}, \bibinfo{person}{Zhi Li},
  \bibinfo{person}{Yucong Luo}, {and} \bibinfo{person}{Enhong Chen}.}
  \bibinfo{year}{2023}\natexlab{b}.
\newblock \showarticletitle{Formertime: Hierarchical multi-scale
  representations for multivariate time series classification}. In
  \bibinfo{booktitle}{\emph{Proceedings of the ACM Web Conference 2023}}.
  \bibinfo{pages}{1437--1445}.
\newblock


\bibitem[Cheng et~al\mbox{.}(2023a)]%
        {cheng2023weakly}
\bibfield{author}{\bibinfo{person}{Yunyao Cheng}, \bibinfo{person}{Peng Chen},
  \bibinfo{person}{Chenjuan Guo}, \bibinfo{person}{Kai Zhao},
  \bibinfo{person}{Qingsong Wen}, \bibinfo{person}{Bin Yang}, {and}
  \bibinfo{person}{Christian~S Jensen}.} \bibinfo{year}{2023}\natexlab{a}.
\newblock \showarticletitle{Weakly guided adaptation for robust time series
  forecasting}.
\newblock \bibinfo{journal}{\emph{Proceedings of the VLDB Endowment}}
  \bibinfo{volume}{17}, \bibinfo{number}{4} (\bibinfo{year}{2023}),
  \bibinfo{pages}{766--779}.
\newblock


\bibitem[Eldele et~al\mbox{.}(2021)]%
        {eldele2021time}
\bibfield{author}{\bibinfo{person}{Emadeldeen Eldele}, \bibinfo{person}{Mohamed
  Ragab}, \bibinfo{person}{Zhenghua Chen}, \bibinfo{person}{Min Wu},
  \bibinfo{person}{Chee~Keong Kwoh}, \bibinfo{person}{Xiaoli Li}, {and}
  \bibinfo{person}{Cuntai Guan}.} \bibinfo{year}{2021}\natexlab{}.
\newblock \showarticletitle{Time-series representation learning via temporal
  and contextual contrasting}.
\newblock \bibinfo{journal}{\emph{arXiv preprint arXiv:2106.14112}}
  (\bibinfo{year}{2021}).
\newblock


\bibitem[Guo et~al\mbox{.}(2023)]%
        {guo2023icassp}
\bibfield{author}{\bibinfo{person}{Hongbo Guo}, \bibinfo{person}{Xinzi Xu},
  \bibinfo{person}{Hao Wu}, {and} \bibinfo{person}{Guoxing Wang}.}
  \bibinfo{year}{2023}\natexlab{}.
\newblock \showarticletitle{Multi-Scale and Multi-Modal Contrastive Learning
  Network for Biomedical Time Series}.
\newblock \bibinfo{journal}{\emph{arXiv preprint arXiv:2312.03796}}
  (\bibinfo{year}{2023}).
\newblock


\bibitem[Guyon and Elisseeff(2003)]%
        {guyon2003introduction}
\bibfield{author}{\bibinfo{person}{Isabelle Guyon} {and}
  \bibinfo{person}{Andr{\'e} Elisseeff}.} \bibinfo{year}{2003}\natexlab{}.
\newblock \showarticletitle{An introduction to variable and feature selection}.
\newblock \bibinfo{journal}{\emph{Journal of machine learning research}}
  \bibinfo{volume}{3}, \bibinfo{number}{Mar} (\bibinfo{year}{2003}),
  \bibinfo{pages}{1157--1182}.
\newblock


\bibitem[Hu et~al\mbox{.}(2020)]%
        {hu2020iterative}
\bibfield{author}{\bibinfo{person}{Ronghang Hu}, \bibinfo{person}{Amanpreet
  Singh}, \bibinfo{person}{Trevor Darrell}, {and} \bibinfo{person}{Marcus
  Rohrbach}.} \bibinfo{year}{2020}\natexlab{}.
\newblock \showarticletitle{Iterative answer prediction with pointer-augmented
  multimodal transformers for textvqa}. In
  \bibinfo{booktitle}{\emph{Proceedings of the IEEE/CVF conference on computer
  vision and pattern recognition}}. \bibinfo{pages}{9992--10002}.
\newblock


\bibitem[Huang et~al\mbox{.}(2023)]%
        {huang2023crossgnn}
\bibfield{author}{\bibinfo{person}{Qihe Huang}, \bibinfo{person}{Lei Shen},
  \bibinfo{person}{Ruixin Zhang}, \bibinfo{person}{Shouhong Ding},
  \bibinfo{person}{Binwu Wang}, \bibinfo{person}{Zhengyang Zhou}, {and}
  \bibinfo{person}{Yang Wang}.} \bibinfo{year}{2023}\natexlab{}.
\newblock \showarticletitle{Crossgnn: Confronting noisy multivariate time
  series via cross interaction refinement}.
\newblock \bibinfo{journal}{\emph{Advances in Neural Information Processing
  Systems}}  \bibinfo{volume}{36} (\bibinfo{year}{2023}),
  \bibinfo{pages}{46885--46902}.
\newblock


\bibitem[Huang et~al\mbox{.}(2025)]%
        {huang2025exploiting}
\bibfield{author}{\bibinfo{person}{Qihe Huang}, \bibinfo{person}{Zhengyang
  Zhou}, \bibinfo{person}{Kuo Yang}, {and} \bibinfo{person}{Yang Wang}.}
  \bibinfo{year}{2025}\natexlab{}.
\newblock \showarticletitle{Exploiting Language Power for Time Series
  Forecasting with Exogenous Variables}. In
  \bibinfo{booktitle}{\emph{Proceedings of the ACM on Web Conference 2025}}.
  \bibinfo{pages}{4043--4052}.
\newblock


\bibitem[Huang et~al\mbox{.}(2019)]%
        {huang2019dsanet}
\bibfield{author}{\bibinfo{person}{Siteng Huang}, \bibinfo{person}{Donglin
  Wang}, \bibinfo{person}{Xuehan Wu}, {and} \bibinfo{person}{Ao Tang}.}
  \bibinfo{year}{2019}\natexlab{}.
\newblock \showarticletitle{Dsanet: Dual self-attention network for
  multivariate time series forecasting}. In
  \bibinfo{booktitle}{\emph{Proceedings of the 28th ACM international
  conference on information and knowledge management}}.
  \bibinfo{pages}{2129--2132}.
\newblock


\bibitem[Jin et~al\mbox{.}(2024)]%
        {jin2024survey}
\bibfield{author}{\bibinfo{person}{Ming Jin}, \bibinfo{person}{Huan~Yee Koh},
  \bibinfo{person}{Qingsong Wen}, \bibinfo{person}{Daniele Zambon},
  \bibinfo{person}{Cesare Alippi}, \bibinfo{person}{Geoffrey~I Webb},
  \bibinfo{person}{Irwin King}, {and} \bibinfo{person}{Shirui Pan}.}
  \bibinfo{year}{2024}\natexlab{}.
\newblock \showarticletitle{A survey on graph neural networks for time series:
  Forecasting, classification, imputation, and anomaly detection}.
\newblock \bibinfo{journal}{\emph{IEEE Transactions on Pattern Analysis and
  Machine Intelligence}} (\bibinfo{year}{2024}).
\newblock


\bibitem[Jin et~al\mbox{.}(2023)]%
        {jin2023time}
\bibfield{author}{\bibinfo{person}{Ming Jin}, \bibinfo{person}{Shiyu Wang},
  \bibinfo{person}{Lintao Ma}, \bibinfo{person}{Zhixuan Chu},
  \bibinfo{person}{James~Y Zhang}, \bibinfo{person}{Xiaoming Shi},
  \bibinfo{person}{Pin-Yu Chen}, \bibinfo{person}{Yuxuan Liang},
  \bibinfo{person}{Yuan-Fang Li}, \bibinfo{person}{Shirui Pan},
  {et~al\mbox{.}}} \bibinfo{year}{2023}\natexlab{}.
\newblock \showarticletitle{Time-llm: Time series forecasting by reprogramming
  large language models}.
\newblock \bibinfo{journal}{\emph{arXiv preprint arXiv:2310.01728}}
  (\bibinfo{year}{2023}).
\newblock


\bibitem[Kieu et~al\mbox{.}(2022)]%
        {kieu2022anomaly}
\bibfield{author}{\bibinfo{person}{Tung Kieu}, \bibinfo{person}{Bin Yang},
  \bibinfo{person}{Chenjuan Guo}, \bibinfo{person}{Razvan-Gabriel Cirstea},
  \bibinfo{person}{Yan Zhao}, \bibinfo{person}{Yale Song}, {and}
  \bibinfo{person}{Christian~S Jensen}.} \bibinfo{year}{2022}\natexlab{}.
\newblock \showarticletitle{Anomaly detection in time series with robust
  variational quasi-recurrent autoencoders}. In \bibinfo{booktitle}{\emph{2022
  IEEE 38th International Conference on Data Engineering (ICDE)}}. IEEE,
  \bibinfo{pages}{1342--1354}.
\newblock


\bibitem[Kingma(2014)]%
        {kingma2014adam}
\bibfield{author}{\bibinfo{person}{Diederik~P Kingma}.}
  \bibinfo{year}{2014}\natexlab{}.
\newblock \showarticletitle{Adam: A method for stochastic optimization}.
\newblock \bibinfo{journal}{\emph{arXiv preprint arXiv:1412.6980}}
  (\bibinfo{year}{2014}).
\newblock


\bibitem[Lai et~al\mbox{.}(2018)]%
        {lai2018lstnet}
\bibfield{author}{\bibinfo{person}{Guokun Lai}, \bibinfo{person}{Wei-Cheng
  Chang}, \bibinfo{person}{Yiming Yang}, {and} \bibinfo{person}{Hanxiao Liu}.}
  \bibinfo{year}{2018}\natexlab{}.
\newblock \showarticletitle{Modeling long-and short-term temporal patterns with
  deep neural networks}. In \bibinfo{booktitle}{\emph{The 41st international
  ACM SIGIR conference on research \& development in information retrieval}}.
  \bibinfo{pages}{95--104}.
\newblock


\bibitem[Lee et~al\mbox{.}(2021)]%
        {lee2021DPT}
\bibfield{author}{\bibinfo{person}{Dongha Lee}, \bibinfo{person}{Seonghyeon
  Lee}, {and} \bibinfo{person}{Hwanjo Yu}.} \bibinfo{year}{2021}\natexlab{}.
\newblock \showarticletitle{Learnable dynamic temporal pooling for time series
  classification}. In \bibinfo{booktitle}{\emph{Proceedings of the AAAI
  Conference on Artificial Intelligence}}, Vol.~\bibinfo{volume}{35}.
  \bibinfo{pages}{8288--8296}.
\newblock


\bibitem[Li et~al\mbox{.}(2021)]%
        {li2021hierarchical}
\bibfield{author}{\bibinfo{person}{Tianfu Li}, \bibinfo{person}{Zhibin Zhao},
  \bibinfo{person}{Chuang Sun}, \bibinfo{person}{Ruqiang Yan}, {and}
  \bibinfo{person}{Xuefeng Chen}.} \bibinfo{year}{2021}\natexlab{}.
\newblock \showarticletitle{Hierarchical attention graph convolutional network
  to fuse multi-sensor signals for remaining useful life prediction}.
\newblock \bibinfo{journal}{\emph{Reliability Engineering \& System Safety}}
  \bibinfo{volume}{215} (\bibinfo{year}{2021}), \bibinfo{pages}{107878}.
\newblock


\bibitem[Lin et~al\mbox{.}(2024)]%
        {lin2024sparsetsf}
\bibfield{author}{\bibinfo{person}{Shengsheng Lin}, \bibinfo{person}{Weiwei
  Lin}, \bibinfo{person}{Wentai Wu}, \bibinfo{person}{Haojun Chen}, {and}
  \bibinfo{person}{Junjie Yang}.} \bibinfo{year}{2024}\natexlab{}.
\newblock \showarticletitle{Sparsetsf: Modeling long-term time series
  forecasting with 1k parameters}.
\newblock \bibinfo{journal}{\emph{arXiv preprint arXiv:2405.00946}}
  (\bibinfo{year}{2024}).
\newblock


\bibitem[Liu et~al\mbox{.}(2025d)]%
        {liu2025st}
\bibfield{author}{\bibinfo{person}{Chenxi Liu},
  \bibinfo{person}{Kethmi~Hirushini Hettige}, \bibinfo{person}{Qianxiong Xu},
  \bibinfo{person}{Cheng Long}, \bibinfo{person}{Shili Xiang},
  \bibinfo{person}{Gao Cong}, \bibinfo{person}{Ziyue Li}, {and}
  \bibinfo{person}{Rui Zhao}.} \bibinfo{year}{2025}\natexlab{d}.
\newblock \showarticletitle{ST-LLM+: Graph Enhanced Spatio-Temporal Large
  Language Models for Traffic Prediction}.
\newblock \bibinfo{journal}{\emph{IEEE Transactions on Knowledge and Data
  Engineering}} (\bibinfo{year}{2025}).
\newblock


\bibitem[Liu et~al\mbox{.}(2025e)]%
        {liu2025timekd}
\bibfield{author}{\bibinfo{person}{Chenxi Liu}, \bibinfo{person}{Hao Miao},
  \bibinfo{person}{Qianxiong Xu}, \bibinfo{person}{Shaowen Zhou},
  \bibinfo{person}{Cheng Long}, \bibinfo{person}{Yan Zhao},
  \bibinfo{person}{Ziyue Li}, {and} \bibinfo{person}{Rui Zhao}.}
  \bibinfo{year}{2025}\natexlab{e}.
\newblock \showarticletitle{Efficient Multivariate Time Series Forecasting via
  Calibrated Language Models with Privileged Knowledge Distillation}. In
  \bibinfo{booktitle}{\emph{41th {IEEE} International Conference on Data
  Engineering}}.
\newblock


\bibitem[Liu et~al\mbox{.}(2025f)]%
        {liu2025timecma}
\bibfield{author}{\bibinfo{person}{Chenxi Liu}, \bibinfo{person}{Qianxiong Xu},
  \bibinfo{person}{Hao Miao}, \bibinfo{person}{Sun Yang},
  \bibinfo{person}{Lingzheng Zhang}, \bibinfo{person}{Cheng Long},
  \bibinfo{person}{Ziyue Li}, {and} \bibinfo{person}{Rui Zhao}.}
  \bibinfo{year}{2025}\natexlab{f}.
\newblock \showarticletitle{Timecma: Towards llm-empowered multivariate time
  series forecasting via cross-modality alignment}. In
  \bibinfo{booktitle}{\emph{Proceedings of the AAAI Conference on Artificial
  Intelligence}}, Vol.~\bibinfo{volume}{39}. \bibinfo{pages}{18780--18788}.
\newblock


\bibitem[Liu et~al\mbox{.}(2024)]%
        {DBLP:journals/corr/abs-2401-10134}
\bibfield{author}{\bibinfo{person}{Chenxi Liu}, \bibinfo{person}{Sun Yang},
  \bibinfo{person}{Qianxiong Xu}, \bibinfo{person}{Zhishuai Li},
  \bibinfo{person}{Cheng Long}, \bibinfo{person}{Ziyue Li}, {and}
  \bibinfo{person}{Rui Zhao}.} \bibinfo{year}{2024}\natexlab{}.
\newblock \showarticletitle{Spatial-Temporal Large Language Model for Traffic
  Prediction}. In \bibinfo{booktitle}{\emph{25th {IEEE} International
  Conference on Mobile Data Management}}. \bibinfo{pages}{31--40}.
\newblock


\bibitem[Liu et~al\mbox{.}(2022b)]%
        {liu2022scinet}
\bibfield{author}{\bibinfo{person}{Minhao Liu}, \bibinfo{person}{Ailing Zeng},
  \bibinfo{person}{Muxi Chen}, \bibinfo{person}{Zhijian Xu},
  \bibinfo{person}{Qiuxia Lai}, \bibinfo{person}{Lingna Ma}, {and}
  \bibinfo{person}{Qiang Xu}.} \bibinfo{year}{2022}\natexlab{b}.
\newblock \showarticletitle{Scinet: Time series modeling and forecasting with
  sample convolution and interaction}.
\newblock \bibinfo{journal}{\emph{Advances in Neural Information Processing
  Systems}}  \bibinfo{volume}{35} (\bibinfo{year}{2022}),
  \bibinfo{pages}{5816--5828}.
\newblock


\bibitem[Liu et~al\mbox{.}(2022a)]%
        {liu2022pyraformer}
\bibfield{author}{\bibinfo{person}{Shizhan Liu}, \bibinfo{person}{Hang Yu},
  \bibinfo{person}{Cong Liao}, \bibinfo{person}{Jianguo Li},
  \bibinfo{person}{Weiyao Lin}, \bibinfo{person}{Alex~X Liu}, {and}
  \bibinfo{person}{Schahram Dustdar}.} \bibinfo{year}{2022}\natexlab{a}.
\newblock \showarticletitle{Pyraformer: Low-complexity pyramidal attention for
  long-range time series modeling and forecasting}. In
  \bibinfo{booktitle}{\emph{\# PLACEHOLDER\_PARENT\_METADATA\_VALUE\#}}.
\newblock


\bibitem[Liu et~al\mbox{.}(2025a)]%
        {liu2025attributed}
\bibfield{author}{\bibinfo{person}{Zhipeng Liu}, \bibinfo{person}{Peibo Duan},
  \bibinfo{person}{Qi Chu}, \bibinfo{person}{Levin Kuhlmann},
  \bibinfo{person}{Changsheng Zhang}, \bibinfo{person}{Wenwei Yue},
  \bibinfo{person}{Xuan Tang}, {and} \bibinfo{person}{Bin Zhang}.}
  \bibinfo{year}{2025}\natexlab{a}.
\newblock \showarticletitle{An Attributed Multiplex Network Enabled GNN-based
  Stock Predictor with Observable and Non-observable Information}.
\newblock \bibinfo{journal}{\emph{Expert Systems with Applications}}
  (\bibinfo{year}{2025}), \bibinfo{pages}{129018}.
\newblock


\bibitem[Liu et~al\mbox{.}(2025b)]%
        {liu2025distillation}
\bibfield{author}{\bibinfo{person}{Zhipeng Liu}, \bibinfo{person}{Peibo Duan},
  \bibinfo{person}{Mingyang Geng}, {and} \bibinfo{person}{Bin Zhang}.}
  \bibinfo{year}{2025}\natexlab{b}.
\newblock \showarticletitle{A Distillation-based Future-aware Graph Neural
  Network for Stock Trend Prediction}. In \bibinfo{booktitle}{\emph{ICASSP
  2025-2025 IEEE International Conference on Acoustics, Speech and Signal
  Processing (ICASSP)}}. IEEE, \bibinfo{pages}{1--5}.
\newblock


\bibitem[Liu et~al\mbox{.}(2025c)]%
        {liu2025dynamic}
\bibfield{author}{\bibinfo{person}{Zhipeng Liu}, \bibinfo{person}{Peibo Duan},
  \bibinfo{person}{Xiaosha Xue}, \bibinfo{person}{Changsheng Zhang},
  \bibinfo{person}{Wenwei Yue}, {and} \bibinfo{person}{Bin Zhang}.}
  \bibinfo{year}{2025}\natexlab{c}.
\newblock \showarticletitle{A dynamic hypergraph attention network: Capturing
  market-wide spatiotemporal dependencies for stock selection}.
\newblock \bibinfo{journal}{\emph{Applied Soft Computing}}
  \bibinfo{volume}{169} (\bibinfo{year}{2025}), \bibinfo{pages}{112524}.
\newblock


\bibitem[Miao et~al\mbox{.}(2025)]%
        {miao2025parameter}
\bibfield{author}{\bibinfo{person}{Hao Miao}, \bibinfo{person}{Ronghui Xu},
  \bibinfo{person}{Yan Zhao}, \bibinfo{person}{Senzhang Wang},
  \bibinfo{person}{Jianxin Wang}, \bibinfo{person}{Philip~S Yu}, {and}
  \bibinfo{person}{Christian~S Jensen}.} \bibinfo{year}{2025}\natexlab{}.
\newblock \showarticletitle{A Parameter-Efficient Federated Framework for
  Streaming Time Series Anomaly Detection via Lightweight Adaptation}.
\newblock \bibinfo{journal}{\emph{TMC}} (\bibinfo{year}{2025}).
\newblock


\bibitem[Miao et~al\mbox{.}(2024)]%
        {miao2024unified}
\bibfield{author}{\bibinfo{person}{Hao Miao}, \bibinfo{person}{Yan Zhao},
  \bibinfo{person}{Chenjuan Guo}, \bibinfo{person}{Bin Yang},
  \bibinfo{person}{Kai Zheng}, \bibinfo{person}{Feiteng Huang},
  \bibinfo{person}{Jiandong Xie}, {and} \bibinfo{person}{Christian~S Jensen}.}
  \bibinfo{year}{2024}\natexlab{}.
\newblock \showarticletitle{A unified replay-based continuous learning
  framework for spatio-temporal prediction on streaming data}. In
  \bibinfo{booktitle}{\emph{2024 IEEE 40th International Conference on Data
  Engineering (ICDE)}}. IEEE, \bibinfo{pages}{1050--1062}.
\newblock


\bibitem[Morid et~al\mbox{.}(2023)]%
        {morid2023time}
\bibfield{author}{\bibinfo{person}{Mohammad~Amin Morid},
  \bibinfo{person}{Olivia R~Liu Sheng}, {and} \bibinfo{person}{Joseph Dunbar}.}
  \bibinfo{year}{2023}\natexlab{}.
\newblock \showarticletitle{Time series prediction using deep learning methods
  in healthcare}.
\newblock \bibinfo{journal}{\emph{ACM Transactions on Management Information
  Systems}} \bibinfo{volume}{14}, \bibinfo{number}{1} (\bibinfo{year}{2023}),
  \bibinfo{pages}{1--29}.
\newblock


\bibitem[Mozer(1991)]%
        {mozer1991induction}
\bibfield{author}{\bibinfo{person}{Michael~C Mozer}.}
  \bibinfo{year}{1991}\natexlab{}.
\newblock \showarticletitle{Induction of multiscale temporal structure}.
\newblock \bibinfo{journal}{\emph{Advances in neural information processing
  systems}}  \bibinfo{volume}{4} (\bibinfo{year}{1991}).
\newblock


\bibitem[Nie et~al\mbox{.}(2022)]%
        {patchTST}
\bibfield{author}{\bibinfo{person}{Yuqi Nie}, \bibinfo{person}{Nam~H Nguyen},
  \bibinfo{person}{Phanwadee Sinthong}, {and} \bibinfo{person}{Jayant
  Kalagnanam}.} \bibinfo{year}{2022}\natexlab{}.
\newblock \showarticletitle{A time series is worth 64 words: Long-term
  forecasting with transformers}.
\newblock \bibinfo{journal}{\emph{arXiv preprint arXiv:2211.14730}}
  (\bibinfo{year}{2022}).
\newblock


\bibitem[Qin et~al\mbox{.}(2017)]%
        {qin2017dual}
\bibfield{author}{\bibinfo{person}{Yao Qin}, \bibinfo{person}{Dongjin Song},
  \bibinfo{person}{Haifeng Chen}, \bibinfo{person}{Wei Cheng},
  \bibinfo{person}{Guofei Jiang}, {and} \bibinfo{person}{Garrison Cottrell}.}
  \bibinfo{year}{2017}\natexlab{}.
\newblock \showarticletitle{A dual-stage attention-based recurrent neural
  network for time series prediction}.
\newblock \bibinfo{journal}{\emph{arXiv preprint arXiv:1704.02971}}
  (\bibinfo{year}{2017}).
\newblock


\bibitem[Qiu et~al\mbox{.}(2024a)]%
        {qiu2024tfb}
\bibfield{author}{\bibinfo{person}{Xiangfei Qiu}, \bibinfo{person}{Jilin Hu},
  \bibinfo{person}{Lekui Zhou}, \bibinfo{person}{Xingjian Wu},
  \bibinfo{person}{Junyang Du}, \bibinfo{person}{Buang Zhang},
  \bibinfo{person}{Chenjuan Guo}, \bibinfo{person}{Aoying Zhou},
  \bibinfo{person}{Christian~S Jensen}, \bibinfo{person}{Zhenli Sheng},
  {et~al\mbox{.}}} \bibinfo{year}{2024}\natexlab{a}.
\newblock \showarticletitle{Tfb: Towards comprehensive and fair benchmarking of
  time series forecasting methods}.
\newblock \bibinfo{journal}{\emph{arXiv preprint arXiv:2403.20150}}
  (\bibinfo{year}{2024}).
\newblock


\bibitem[Qiu et~al\mbox{.}(2024b)]%
        {qiu2024duet}
\bibfield{author}{\bibinfo{person}{Xiangfei Qiu}, \bibinfo{person}{Xingjian
  Wu}, \bibinfo{person}{Yan Lin}, \bibinfo{person}{Chenjuan Guo},
  \bibinfo{person}{Jilin Hu}, {and} \bibinfo{person}{Bin Yang}.}
  \bibinfo{year}{2024}\natexlab{b}.
\newblock \showarticletitle{Duet: Dual clustering enhanced multivariate time
  series forecasting}.
\newblock \bibinfo{journal}{\emph{arXiv preprint arXiv:2412.10859}}
  (\bibinfo{year}{2024}).
\newblock


\bibitem[Sen et~al\mbox{.}(2019)]%
        {sen2019think}
\bibfield{author}{\bibinfo{person}{Rajat Sen}, \bibinfo{person}{Hsiang-Fu Yu},
  {and} \bibinfo{person}{Inderjit~S Dhillon}.} \bibinfo{year}{2019}\natexlab{}.
\newblock \showarticletitle{Think globally, act locally: A deep neural network
  approach to high-dimensional time series forecasting}.
\newblock \bibinfo{journal}{\emph{Advances in neural information processing
  systems}}  \bibinfo{volume}{32} (\bibinfo{year}{2019}).
\newblock


\bibitem[Stankeviciute et~al\mbox{.}(2021)]%
        {stankeviciute2021conformal}
\bibfield{author}{\bibinfo{person}{Kamile Stankeviciute},
  \bibinfo{person}{Ahmed M~Alaa}, {and} \bibinfo{person}{Mihaela van~der
  Schaar}.} \bibinfo{year}{2021}\natexlab{}.
\newblock \showarticletitle{Conformal time-series forecasting}.
\newblock \bibinfo{journal}{\emph{Advances in neural information processing
  systems}}  \bibinfo{volume}{34} (\bibinfo{year}{2021}),
  \bibinfo{pages}{6216--6228}.
\newblock


\bibitem[Subramanian et~al\mbox{.}(2020)]%
        {subramanian2020multi}
\bibfield{author}{\bibinfo{person}{Sandeep Subramanian}, \bibinfo{person}{Ronan
  Collobert}, \bibinfo{person}{Marc'Aurelio Ranzato}, {and}
  \bibinfo{person}{Y-Lan Boureau}.} \bibinfo{year}{2020}\natexlab{}.
\newblock \showarticletitle{Multi-scale Transformer Language Models}.
\newblock \bibinfo{journal}{\emph{arXiv preprint arXiv:2005.00581}}
  (\bibinfo{year}{2020}).
\newblock


\bibitem[Teng et~al\mbox{.}(2022)]%
        {teng_2022_multistock}
\bibfield{author}{\bibinfo{person}{Xiao Teng}, \bibinfo{person}{Xiang Zhang},
  {and} \bibinfo{person}{Zhigang Luo}.} \bibinfo{year}{2022}\natexlab{}.
\newblock \showarticletitle{Multi-scale local cues and hierarchical
  attention-based LSTM for stock price trend prediction}.
\newblock \bibinfo{journal}{\emph{Neurocomputing}}  \bibinfo{volume}{505}
  (\bibinfo{year}{2022}), \bibinfo{pages}{92--100}.
\newblock


\bibitem[Wang et~al\mbox{.}(2023b)]%
        {wang2023pattern}
\bibfield{author}{\bibinfo{person}{Binwu Wang}, \bibinfo{person}{Yudong Zhang},
  \bibinfo{person}{Xu Wang}, \bibinfo{person}{Pengkun Wang},
  \bibinfo{person}{Zhengyang Zhou}, \bibinfo{person}{Lei Bai}, {and}
  \bibinfo{person}{Yang Wang}.} \bibinfo{year}{2023}\natexlab{b}.
\newblock \showarticletitle{Pattern expansion and consolidation on evolving
  graphs for continual traffic prediction}. In
  \bibinfo{booktitle}{\emph{Proceedings of the 29th ACM SIGKDD Conference on
  Knowledge Discovery and Data Mining}}. \bibinfo{pages}{2223--2232}.
\newblock


\bibitem[Wang et~al\mbox{.}(2023c)]%
        {wang2023drift}
\bibfield{author}{\bibinfo{person}{Chengsen Wang}, \bibinfo{person}{Zirui
  Zhuang}, \bibinfo{person}{Qi Qi}, \bibinfo{person}{Jingyu Wang},
  \bibinfo{person}{Xingyu Wang}, \bibinfo{person}{Haifeng Sun}, {and}
  \bibinfo{person}{Jianxin Liao}.} \bibinfo{year}{2023}\natexlab{c}.
\newblock \showarticletitle{Drift doesn't matter: dynamic decomposition with
  diffusion reconstruction for unstable multivariate time series anomaly
  detection}.
\newblock \bibinfo{journal}{\emph{Advances in Neural Information Processing
  Systems}}  \bibinfo{volume}{36} (\bibinfo{year}{2023}),
  \bibinfo{pages}{10758--10774}.
\newblock


\bibitem[Wang et~al\mbox{.}(2023a)]%
        {MICN}
\bibfield{author}{\bibinfo{person}{Huiqiang Wang}, \bibinfo{person}{Jian Peng},
  \bibinfo{person}{Feihu Huang}, \bibinfo{person}{Jince Wang},
  \bibinfo{person}{Junhui Chen}, {and} \bibinfo{person}{Yifei Xiao}.}
  \bibinfo{year}{2023}\natexlab{a}.
\newblock \showarticletitle{Micn: Multi-scale local and global context modeling
  for long-term series forecasting}. In \bibinfo{booktitle}{\emph{The eleventh
  international conference on learning representations}}.
\newblock


\bibitem[Wang et~al\mbox{.}(2024b)]%
        {wang_2024_Timemixer}
\bibfield{author}{\bibinfo{person}{Shiyu Wang}, \bibinfo{person}{Haixu Wu},
  \bibinfo{person}{Xiaoming Shi}, \bibinfo{person}{Tengge Hu},
  \bibinfo{person}{Huakun Luo}, \bibinfo{person}{Lintao Ma},
  \bibinfo{person}{James~Y Zhang}, {and} \bibinfo{person}{Jun Zhou}.}
  \bibinfo{year}{2024}\natexlab{b}.
\newblock \showarticletitle{Timemixer: Decomposable multiscale mixing for time
  series forecasting}.
\newblock \bibinfo{journal}{\emph{arXiv preprint arXiv:2405.14616}}
  (\bibinfo{year}{2024}).
\newblock


\bibitem[Wang et~al\mbox{.}(2021)]%
        {wang2021pyramid}
\bibfield{author}{\bibinfo{person}{Wenhai Wang}, \bibinfo{person}{Enze Xie},
  \bibinfo{person}{Xiang Li}, \bibinfo{person}{Deng-Ping Fan},
  \bibinfo{person}{Kaitao Song}, \bibinfo{person}{Ding Liang},
  \bibinfo{person}{Tong Lu}, \bibinfo{person}{Ping Luo}, {and}
  \bibinfo{person}{Ling Shao}.} \bibinfo{year}{2021}\natexlab{}.
\newblock \showarticletitle{Pyramid vision transformer: A versatile backbone
  for dense prediction without convolutions}. In
  \bibinfo{booktitle}{\emph{Proceedings of the IEEE/CVF international
  conference on computer vision}}. \bibinfo{pages}{568--578}.
\newblock


\bibitem[Wang et~al\mbox{.}(2024a)]%
        {wang2024survey}
\bibfield{author}{\bibinfo{person}{Yuxuan Wang}, \bibinfo{person}{Haixu Wu},
  \bibinfo{person}{Jiaxiang Dong}, \bibinfo{person}{Yong Liu},
  \bibinfo{person}{Mingsheng Long}, {and} \bibinfo{person}{Jianmin Wang}.}
  \bibinfo{year}{2024}\natexlab{a}.
\newblock \showarticletitle{Deep time series models: A comprehensive survey and
  benchmark}.
\newblock \bibinfo{journal}{\emph{arXiv preprint arXiv:2407.13278}}
  (\bibinfo{year}{2024}).
\newblock


\bibitem[Wang et~al\mbox{.}(2024c)]%
        {wang2024graph}
\bibfield{author}{\bibinfo{person}{Yucheng Wang}, \bibinfo{person}{Yuecong Xu},
  \bibinfo{person}{Jianfei Yang}, \bibinfo{person}{Min Wu},
  \bibinfo{person}{Xiaoli Li}, \bibinfo{person}{Lihua Xie}, {and}
  \bibinfo{person}{Zhenghua Chen}.} \bibinfo{year}{2024}\natexlab{c}.
\newblock \showarticletitle{Graph-Aware Contrasting for Multivariate
  Time-Series Classification}. In \bibinfo{booktitle}{\emph{Proceedings of the
  AAAI Conference on Artificial Intelligence}}, Vol.~\bibinfo{volume}{38}.
  \bibinfo{pages}{15725--15734}.
\newblock


\bibitem[Wu et~al\mbox{.}(2021)]%
        {wu2021autoformer}
\bibfield{author}{\bibinfo{person}{Haixu Wu}, \bibinfo{person}{Jiehui Xu},
  \bibinfo{person}{Jianmin Wang}, {and} \bibinfo{person}{Mingsheng Long}.}
  \bibinfo{year}{2021}\natexlab{}.
\newblock \showarticletitle{Autoformer: Decomposition transformers with
  auto-correlation for long-term series forecasting}.
\newblock \bibinfo{journal}{\emph{Advances in neural information processing
  systems}}  \bibinfo{volume}{34} (\bibinfo{year}{2021}),
  \bibinfo{pages}{22419--22430}.
\newblock


\bibitem[Ye et~al\mbox{.}(2025)]%
        {ye2025non}
\bibfield{author}{\bibinfo{person}{Weiwei Ye}, \bibinfo{person}{Zhuopeng Xu},
  {and} \bibinfo{person}{Ning Gui}.} \bibinfo{year}{2025}\natexlab{}.
\newblock \showarticletitle{Non-stationary Diffusion For Probabilistic Time
  Series Forecasting}.
\newblock \bibinfo{journal}{\emph{arXiv preprint arXiv:2505.04278}}
  (\bibinfo{year}{2025}).
\newblock


\bibitem[Zhang et~al\mbox{.}(2025)]%
        {zhang2025strap}
\bibfield{author}{\bibinfo{person}{Haoyu Zhang}, \bibinfo{person}{Wentao
  Zhang}, \bibinfo{person}{Hao Miao}, \bibinfo{person}{Xinke Jiang},
  \bibinfo{person}{Yuchen Fang}, {and} \bibinfo{person}{Yifan Zhang}.}
  \bibinfo{year}{2025}\natexlab{}.
\newblock \showarticletitle{STRAP: Spatio-Temporal Pattern Retrieval for
  Out-of-Distribution Generalization}.
\newblock \bibinfo{journal}{\emph{arXiv preprint arXiv:2505.19547}}
  (\bibinfo{year}{2025}).
\newblock


\bibitem[Zhang and Yan(2023)]%
        {zhang2023crossformer}
\bibfield{author}{\bibinfo{person}{Yunhao Zhang} {and} \bibinfo{person}{Junchi
  Yan}.} \bibinfo{year}{2023}\natexlab{}.
\newblock \showarticletitle{Crossformer: Transformer utilizing cross-dimension
  dependency for multivariate time series forecasting}. In
  \bibinfo{booktitle}{\emph{The eleventh international conference on learning
  representations}}.
\newblock


\bibitem[Zhong et~al\mbox{.}(2023)]%
        {zhong2023mlpmixer}
\bibfield{author}{\bibinfo{person}{Shuhan Zhong}, \bibinfo{person}{Sizhe Song},
  \bibinfo{person}{Weipeng Zhuo}, \bibinfo{person}{Guanyao Li},
  \bibinfo{person}{Yang Liu}, {and} \bibinfo{person}{S-H~Gary Chan}.}
  \bibinfo{year}{2023}\natexlab{}.
\newblock \showarticletitle{A multi-scale decomposition mlp-mixer for time
  series analysis}.
\newblock \bibinfo{journal}{\emph{arXiv preprint arXiv:2310.11959}}
  (\bibinfo{year}{2023}).
\newblock


\bibitem[Zhou et~al\mbox{.}(2022)]%
        {zhou2022fedformer}
\bibfield{author}{\bibinfo{person}{Tian Zhou}, \bibinfo{person}{Ziqing Ma},
  \bibinfo{person}{Qingsong Wen}, \bibinfo{person}{Xue Wang},
  \bibinfo{person}{Liang Sun}, {and} \bibinfo{person}{Rong Jin}.}
  \bibinfo{year}{2022}\natexlab{}.
\newblock \showarticletitle{Fedformer: Frequency enhanced decomposed
  transformer for long-term series forecasting}. In
  \bibinfo{booktitle}{\emph{International conference on machine learning}}.
  PMLR, \bibinfo{pages}{27268--27286}.
\newblock


\bibitem[Zhou et~al\mbox{.}(2023)]%
        {zhou2023one}
\bibfield{author}{\bibinfo{person}{Tian Zhou}, \bibinfo{person}{Peisong Niu},
  \bibinfo{person}{Liang Sun}, \bibinfo{person}{Rong Jin}, {et~al\mbox{.}}}
  \bibinfo{year}{2023}\natexlab{}.
\newblock \showarticletitle{One fits all: Power general time series analysis by
  pretrained lm}.
\newblock \bibinfo{journal}{\emph{Advances in neural information processing
  systems}}  \bibinfo{volume}{36} (\bibinfo{year}{2023}),
  \bibinfo{pages}{43322--43355}.
\newblock


\bibitem[Zhu et~al\mbox{.}(2023)]%
        {zhu2023long}
\bibfield{author}{\bibinfo{person}{Chenglong Zhu}, \bibinfo{person}{Xueling
  Ma}, \bibinfo{person}{Weiping Ding}, {and} \bibinfo{person}{Jianming Zhan}.}
  \bibinfo{year}{2023}\natexlab{}.
\newblock \showarticletitle{Long-term time series forecasting with multi-linear
  trend fuzzy information granules for LSTM in a periodic framework}.
\newblock \bibinfo{journal}{\emph{IEEE Transactions on Fuzzy Systems}}
  (\bibinfo{year}{2023}).
\newblock


\end{thebibliography}


\end{document}